\documentclass[review,12pt,3p]{elsarticle}
\let\today\relax
\makeatletter
\def\ps@pprintTitle{%
    \let\@oddhead\@empty
    \let\@evenhead\@empty
    \def\@oddfoot{\footnotesize\itshape
         {} \hfill\today}%
    \let\@evenfoot\@oddfoot
    }
\makeatother

\usepackage{algpseudocode}
\usepackage{algorithm}
\usepackage[toc,page]{appendix}
\usepackage{lineno}
\usepackage{ulem}

\usepackage{graphics}
\usepackage{graphicx}
\usepackage[numbers]{natbib}
\modulolinenumbers[5]
\usepackage[colorlinks=true,draft=false]{hyperref}
\usepackage{subcaption}
\usepackage{subeqnarray}
\usepackage{amsmath}
\usepackage{amsfonts}
\usepackage{xcolor,colortbl}
\usepackage[nameinlink, capitalize]{cleveref}
\usepackage{lipsum}
\usepackage{mwe}
\usepackage{rotating}
\usepackage{siunitx}

\newcommand\longvar[1]{\mathchardef\UrlBreakPenalty=100
\mathchardef\UrlBigBreakPenalty=100\nolinkurl{#1}}
\newcommand{\mytilde}{\raise.17ex\hbox{$\scriptstyle\mathtt{\sim}$}}

\usepackage{tikz}

\begin{document}
\begin{frontmatter}
\title{Recurrent Convolutional Deep Neural Networks for Modeling Time-resolved Wildfire Spread Behavior}

\author[GOOGLE]{John Burge\corref{cor1}}
\ead{lawnguy@google.com}
\author[STANFORD]{Matthew R. Bonanni}
\author[GOOGLE]{R. Lily Hu}
\author[GOOGLE,STANFORD,SLAC]{Matthias Ihme}

\address[GOOGLE]{Google Research, Mountain View, CA 94043, USA}
\address[STANFORD]{Department of Mechanical Engineering, Stanford University, CA 94305, USA}
\address[SLAC]{Department of Photon Science, SLAC National Accelerator Laboratory, Menlo Park, CA 94025, USA}

\begin{abstract}
The increasing incidence and severity of wildfires underscores the necessity of accurately predicting their behavior. While high-fidelity models derived from first principles offer physical accuracy, they are too computationally expensive for use in real-time fire response. Low-fidelity models sacrifice some physical accuracy and generalizability via the integration of empirical measurements, but enable real-time simulations for operational use in fire response. Machine learning techniques offer the ability to bridge these objectives by learning first-principles physics while achieving computational speedup. While deep learning approaches have demonstrated the ability to predict wildfire propagation over large time periods, time-resolved fire-spread predictions are needed for active fire management. In this work, we evaluate the ability of  deep learning approaches in accurately modeling the time-resolved dynamics of wildfires.  We use an autoregressive process in which a convolutional recurrent deep learning model makes predictions that propagate a wildfire over 15 minute increments. We demonstrate the model in application to three simulated datasets of increasing complexity, containing both field fires with homogeneous fuel distribution as well as real-world topologies sampled from the California region of the United States.  We show that even after 100 autoregressive predictions representing more than 24 hours of simulated fire spread, the resulting models generate stable and realistic propagation dynamics, achieving a Jaccard score between 0.89 and 0.94 when predicting the resulting fire scar.
\end{abstract}


\end{frontmatter}
%%%%%%%%%%%%%%%%%%%%%%%%%%%
\tableofcontents
%%%%%%%%%%%%%%%%%%%%%%%%%%%
\section{Introduction}
%%%%%%%%%%%%%%%%%%%%%%%%%%%

In recent decades, climate change and excessive fire suppression have resulted in an increase in both the size and severity of wildfires~\cite{covington_southwestern_1994,westerling_warming_2006,mueller_climate_2020}. In addition to the impact of wildfires on 
property damage and loss of life~\cite{thomas_costs_2017}, they also have long-term effects on both health~\cite{kollanus_mortality_2017} and the environment~\cite{van_der_werf_global_2017}. Understanding and modeling wildfire spread is critical to informing direct response to ongoing fires and is an important tool for evaluating fire risk.

Computational models predicting wildfire behavior for fire management, risk assessment, and wildfire mitigation have seen significant advancements \cite{BAKHSHAII_JOHNSON_ETAL_CJFR2019}. These models can be categorized as physical, empirical, and mathematical analogues \cite{Sullivan2009a,Sullivan2009b,Sullivan2009c}. Physical models are built from first principles and provide a high level of fidelity in predicting the fire behavior by solving conservation equations to capture relevant physical processes. Notable examples of these include FIRETEC~\cite{Linn1997}, IUSTI~\cite{larini_multiphase_1998}, and WFDS \cite{mell_physics-based_2007}. The computational complexity of these models currently prevents their application to real-time simulations, so that they are often used in augmenting field experimentation or in the detailed analysis of wildfire dynamics~\cite{Sullivan2009a}.

In contrast, purely empirical models rely solely on data, making no assumptions about fire behavior based on theory \cite{Sullivan2009b}.  Quasi-empirical models combine observations from real fires and laboratory experiments with knowledge of the underlying combustion and heat-transfer processes. One such model is Rothermel's formulation~\cite{Rothermel1972}, which serves as cornerstone for several fire prediction methods used in the United States, such as BEHAVE~\cite{andrews_behave_1986} and FARSITE~\cite{Finney1998}. While these models offer real-time simulations, enabling their use in operational settings, their development involves a manual process of model tuning, based on current understanding of wildland fire spread. 

In the past decade, the potential of machine learning (ML) methods for application to wildfires has been recognized, with ML techniques used across tasks such as fuel characterization, risk assessment, fire behavior modeling, and fire management~\cite{JAIN_ETAL_ER2020,IHME_CHUNG_MISHRA_PECS2022}.  While there are various approaches within the ML community, the sub-field of deep learning with the development of deep neural network (DNN) has led to breakthroughs in many domains~\cite{LECUN_ETAL_N2015}.  However, only a few studies have examined the utility of deep learning methods for predicting the dynamics of wildfire propagation~\cite{Hodges2019,firecast,Burge2020,Bolt2022}. \citet{Hodges2019} demonstrated that a specific DNN, the Deep Converse Inverse Graphical Model (DCIGN) \cite{Kulkarni2015}, could effectively predict the state of a fire six hours into the future in a single time step.  They used FARSITE to generate a database to train the model, and this single six-hour prediction could be computed orders of magnitude faster than explicitly time-advancing the FARSITE simulation. \citet{firecast} performed a similar study by modeling the 24-hour evolution of fire propagating over the geography of the Rocky Mountain region of the United States using real-world data instead of simulated data.

While performing a single 6- or 24-hour prediction with a DNN was shown to be much faster than running FARSITE over the required number of simulation steps, the lack of temporally resolving the wildfire dynamics limits the application for active fire interventions and fire management.  By addressing this issue, our objective is to examine the utility of DNNs for modeling the wildfire propagation dynamics on fine temporal scales with time increments of minutes. To provide predictions over larger time frames, we use an autoregressive process where the output for the prediction at time $\tau$ is used as input for the prediction at time $\tau+\Delta\tau$, where $\Delta\tau=15$ minutes.

We show that when dealing with temporal dynamics, the DCIGN model is not capable of accurately replicating propagation beyond, at most, ten autoregressive predictions before failing to make any realistic predictions at all. To address this issue, we develop an Encoder-Processor-Decoder (EPD) model~\cite{EPD,EPD_stephan} to represent the spatial relationships in the data and add recurrent transformations to represent temporal relationships that were not considered in prior work. Our model, the EPD-ConvLSTM model, is summarized in~\cref{sec:Models}, and the process for training and evaluating the models are presented in~\cref{sec:Methods}.  Results are discussed in~\cref{sec:Results}, and the manuscript closes in~\cref{sec:Conclusion} with conclusions and potential future research directions.

%%%%%%%%%%%%%%%%%%%%%%%%%%%
\section{\label{sec:Data}Generation of Dataset}
%%%%%%%%%%%%%%%%%%%%%%%%%%%

All DNN models were trained and evaluated using synthetic datasets created from FARSITE~\cite{Finney1998}. A total of 30,000 fire simulations were generated across three datasets. Each of these used a domain that was discretized by $128\times 128$ cells with a cell size of 30 m.

The ranges for parameters and operating conditions used to generate the dataset are summarized in~\cref{tab:parameters}. All parameters are held constant over time. Fuel types across all three datasets were chosen from the dynamic fuel models defined by~\citet{Scott2005}, without considering ground fuels. The method introduced by \citet{Finney1998} was applied for calculating the crown fire potential, with a specified foliar moisture content of 100\%. Spotting was not considered. The starting location for each fire was randomly selected to be within the central 50\% of the field.  An octagonal fire front with a width of approximately 75 m (\mytilde2\% of the width of the field) was then placed at this location, and allowed to propagate outwards. The duration of each simulation was held constant at 72 hours with a time step size of 15 minutes, yielding sequences with 289 time steps. The effects of varying solar radiation due to the earth's rotation were not considered, with the latitude, longitude, and elevation of each landscape set identically for each simulation to $0^\circ$, $0^\circ$, and $0\,\si{m}$, respectively. The sun's position in the sky was held constant at $90^\circ$ of elevation and $0^\circ$ of azimuth. Finally, in order the eliminate boundary effects, one cell was trimmed from the border, resulting in a final field size of $126\times 126$ cells.
%==================
\begin{table}[htb!]
\centering
\caption{Parameter ranges used for generating synthetic datasets from FARSITE.}
\label{tab:parameters}
\begin{tabular}{|l|c|}
\hline
Parameter [Units]   & Range \\
\hline\hline
Slope [$^\circ$]                & $[0,45]$        \\
Aspect [$^\circ$]               & $[0, 360]$\\
Wind direction [$^\circ$]       & $[0, 360]$\\
Wind velocity [km/h]            & $[0,50]$\\
Fuel model [\#]                 & $[1,40]$\\
1-hour moisture [\%]            &  $[2,40]$\\
10-hour moisture [\%]           &  $[2,40]$\\
100-hour moisture [\%]          &  $[2,40]$\\
Live herbaceous moisture [\%]   & $[30,100]$\\
Live woody moisture [\%]        & $[30,100]$\\
Canopy cover [\%]               & $[0,100]$\\
Canopy height [m]               & $[3,50]$\\
Crown ratio [--]                & $[0.1,1]$\\
Canopy bulk density [kg/m$^3$]  & $[0,4000]$\\
\hline
\end{tabular}
\end{table}
%==================

The output of each simulation is a series of data points for the position of the fire front as it advances over time. To transform the results into gridded data for training ML models, the burned area for each cell was computed as the fraction of the burned region inside each cell.

Based on these simulations, a dataset was constructed with 17 channels, where each channel represents a specific parameter. The state of the fire at a fixed point in time is represented by a three dimensional tensor: $X\in\mathbb{R}^{H \times W \times C}$, where $H$ is the height, $W$ is the width, and $C$ the number of channels; here $H=W=126$ and $C=17$.  The first three channels were derived from the burn fraction described above: (1) \longvar{vegetation}, the fraction of vegetation that remains unburnt, (2) \longvar{fire\_front}, the fraction of vegetation that burned in the previous time step, and (3) \longvar{scar}, the fraction of vegetation that remains unburnt.  The remaining 14 channels were taken directly from the following field variables tracked in FARSITE~\cite{Finney1998}: (4) \longvar{wind\_east}, (5) \longvar{wind\_north}, (6) \longvar{moisture\_1\_hour}, (7) \longvar{moisture\_10\_hour}, (8) \longvar{moisture\_100\_hour}, (9) \longvar{moisture\_live\_herbaceous}, (10) \longvar{moisture\_live\_woody}, (11) \longvar{cover}, (12) \longvar{height}, (13) \longvar{base}, (14) \longvar{density}, (15) \longvar{slope\_east}, (16) \longvar{slope\_north},
and (17) \longvar{fuel}.  All channels contain continuous data, with the sole exception of the fuel channel, which contains a discrete index that  specifies the fuel model.  

Three datasets with varying complexity and characteristics were designed to evaluate specific components of the ML models.

\begin{itemize}
    \item \textit{single fuel} -- Every simulation used the same fuel model across the entire landscape: GR1, or ``short, sparse dry climate grass". All other parameters were held uniform across a given landscape, but vary randomly across individual simulations, and were selected from a uniform distribution within the ranges described in \cref{tab:parameters}. The terrain is planar, with a constant slope and aspect. This dataset was designed to evaluate the ML models' ability to respond to each of the changing parameters without spatial variation.
    \item \textit{multiple fuel} -- Generated identically to \textit{single fuel}, with the addition that for each fire sequence, a random fuel model is used for the entire domain.  This seemingly minor deviation from the \textit{single fuel} dataset results in approximately 40 times less coverage of the possible combinations of wind magnitudes, terrain slopes and fuel types, the three primary drivers of propagation dynamics, and thus evaluates the ability for models to generalize effectively.
    \item \textit{California} -- Real-world Landscape (LCP) 40 data was obtained from the LANDFIRE program \cite{Rollins2009} for the continental United States, which was then bounded by the north-south and east-west extents of the state of California and includes a significant portion of Nevada as well. For each simulation, a random ($128\times 128$) field was selected from the region. In order to avoid large nonburnable regions such as bodies of water, if the field did not consist of at least 70\% burnable fuel, it was resampled. Wind data was randomly generated in the same manner as in the other two datasets.
\end{itemize}

%===================
\begin{figure}[htb!]
    \centering
    \includegraphics[width=0.95\textwidth]{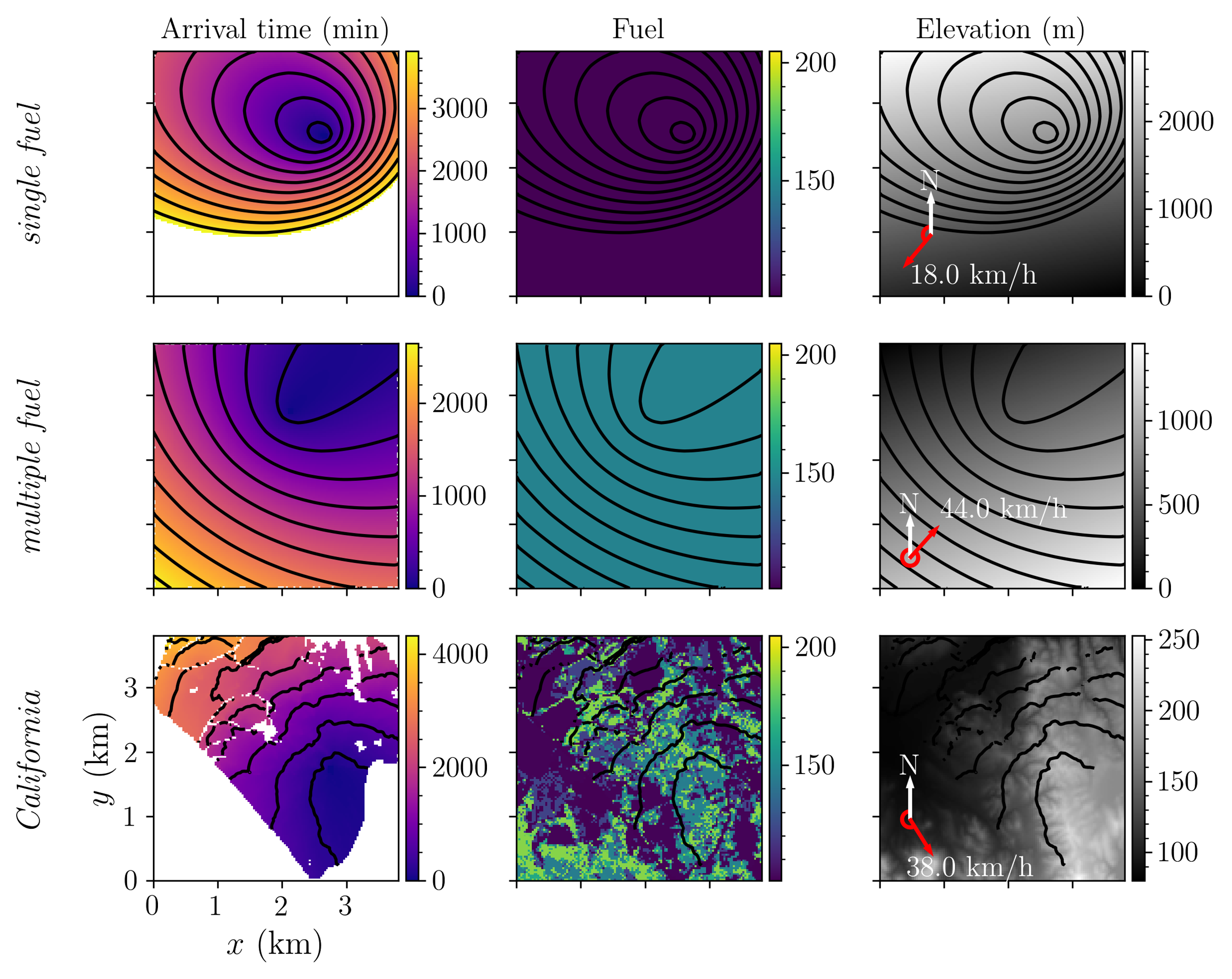}
    \caption{\label{fig:dataset_samples}Sample results from the three datasets used for training the ML models. Isocontours of arrival time are overlaid in black.}
\end{figure}
%===================

A representative sample from each of the three datasets is shown in~\cref{fig:dataset_samples}. In the sample taken from the \textit{single fuel} dataset, we note the expected elliptical fire spread pattern, where the direction is influenced by both the slope of the terrain and the presence of the wind. In the \textit{multiple fuel} dataset, while the fuel is still homogeneous, a different fuel has been selected when compared to the first sample. Despite the significantly higher wind speed, the fire has a much faster upwind spread rate, demonstrating the significantly different properties of this fuel. Finally, the \textit{California} dataset demonstrates all of the dynamics present in this complex dataset. The fire front is seen responding to different fuels by spreading at different rates, and is even blocked completely in some regions, including a road crossing the terrain in the southwest corner. The primary direction of spread is influenced by the wind, while we also see modulations associated with the terrain, where steep downward slopes impede the rate of spread.

These simulations were performed in parallel on a high-performance computing cluster, with each simulation taking an average of 30 seconds to run single-threaded on 1 CPU, for a total cost of 250 CPU-hours, or 9,000 core-hours on 36-core processors. The post-processing of the simulations to compute the fractional burned area of each cell as described above consumed an additional 13,000 core-hours, for a total of approximately 21,000 core-hours.

%%%%%%%%%%%%%%%%%%%%
\section{\label{sec:Models}Machine-learning Models}
%%%%%%%%%%%%%%%%%%%%
%\subsection{\label{sec:deep_neural_networks}Deep Neural Networks}
%%%%%%%%%%%%%%%%%%%%
In the present work, we will use DNNs to model the fire-spread behavior. A detailed discussion of DNNs is beyond the scope of this paper, but for readers unfamiliar with the commonly employed transformations used in our DNNs, we provide a summary in~\ref{appendix:brief dnn background}.  For more detail, we refer the interested reader to the text by~\citet{GenerativeDeepLearning}.

In the remainder of this section, we describe the structure of the DCIGN model, the EPD model and the EPD-ConvLSTM model.  We represent the input to the model as $X$.  For models that need just a single time point as input, $X\in\mathbb{R}^{B \times H \times W \times C}$, where $B$ is the size of the training batch.  For models that need temporal input, $X\in\mathbb{R}^{B \times T \times H \times W \times C}$, where $T$ is the length of the time series.

%%%%%%%%%%%%%%%%%%%%%
\subsection{\label{sec:DCIGN}DCIGN Model}
%%%%%%%%%%%%%%%%%%%%%

\begin{figure}[!htb!]
    \centering
    \includegraphics[width=0.95\textwidth]{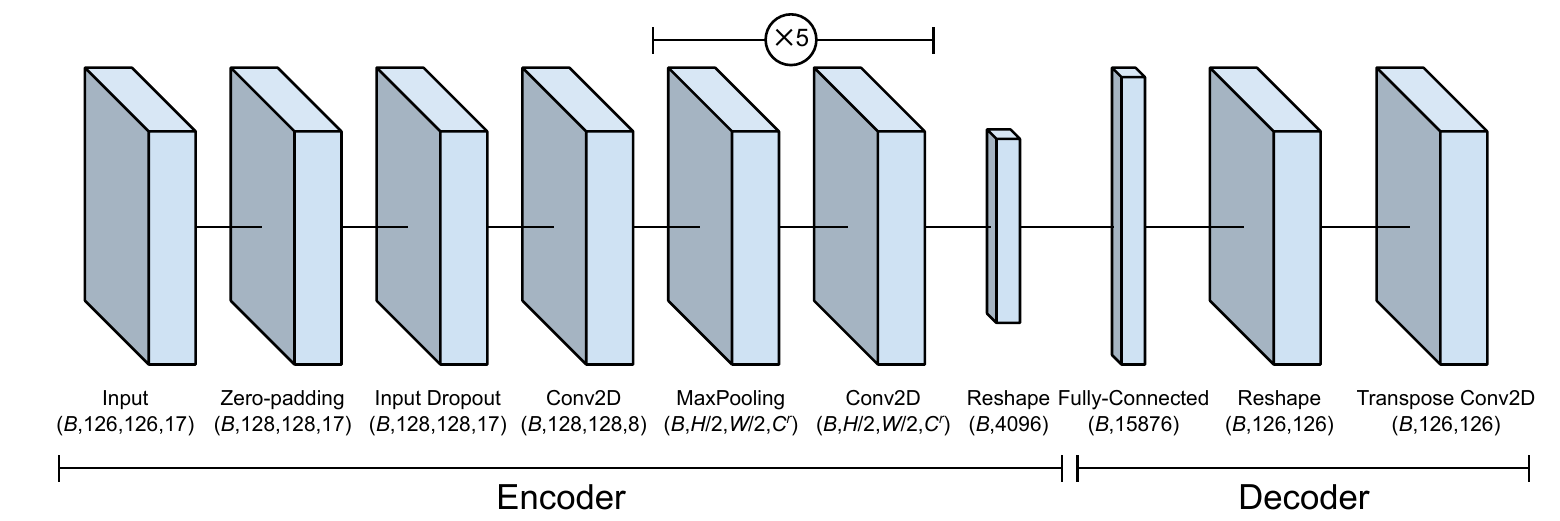}
    \caption{\label{fig:dcign}Schematic of DCIGN architecture employed in the present work.  Parenthesis provide the shape of the data produced by the corresponding layer in the model.}
\end{figure}

Many DNNs can be broken up into two distinct stages.  First, an encoder transforms the input into a latent space and second, a decoder transforms the latent space into the desired output. The DCIGN model is an instance of such a model, \cref{fig:dcign}.  The encoder starts by adding some zero-padding to get the spatial dimensions to be powers of 2.  Then, a repeated set of convolutional transformations and downsampling steps are applied. This ultimately transforms the input of shape $(B,H=126,W=126,C=17)$ into the latent space of shape $(B,H=8,W=8,C=64)$.  In essence, the convolutional transformations have moved the input with 17 distinct channels into a new latent space that has 64 distinct channels but a much smaller spatial resolution.  The decoder transforms this latent space by using a fully connected layer into 15,876 scalar values, which are subsequently reshaped into a $(B,H=126,W=126)$ field that is the output of the DNN model.

%====================
%====================

The hyperparameters for the model were selected to be as close to the work of \citet{Hodges2019} as possible.  Thus, the convolutional kernels had the size (10,10) and the number of filters for the five repeated convolutions were set to: 8, 8, 16, 32 and 64 (i.e., these are the values for $C^r$ in the shape of the convolutions in~\cref{fig:dcign}).  The ideal learning rate was found to be $10^{-4}$.  Their model included dropout layers to address overfitting, which was unnecessary here as we used approximately 100 times more training samples.

%%%%%%%%%%%%%%%%%%%%%
\subsection{\label{sec:EPD}EPD Model}
\begin{figure}[htb!]
    \centering
    \includegraphics[width=0.95\textwidth]{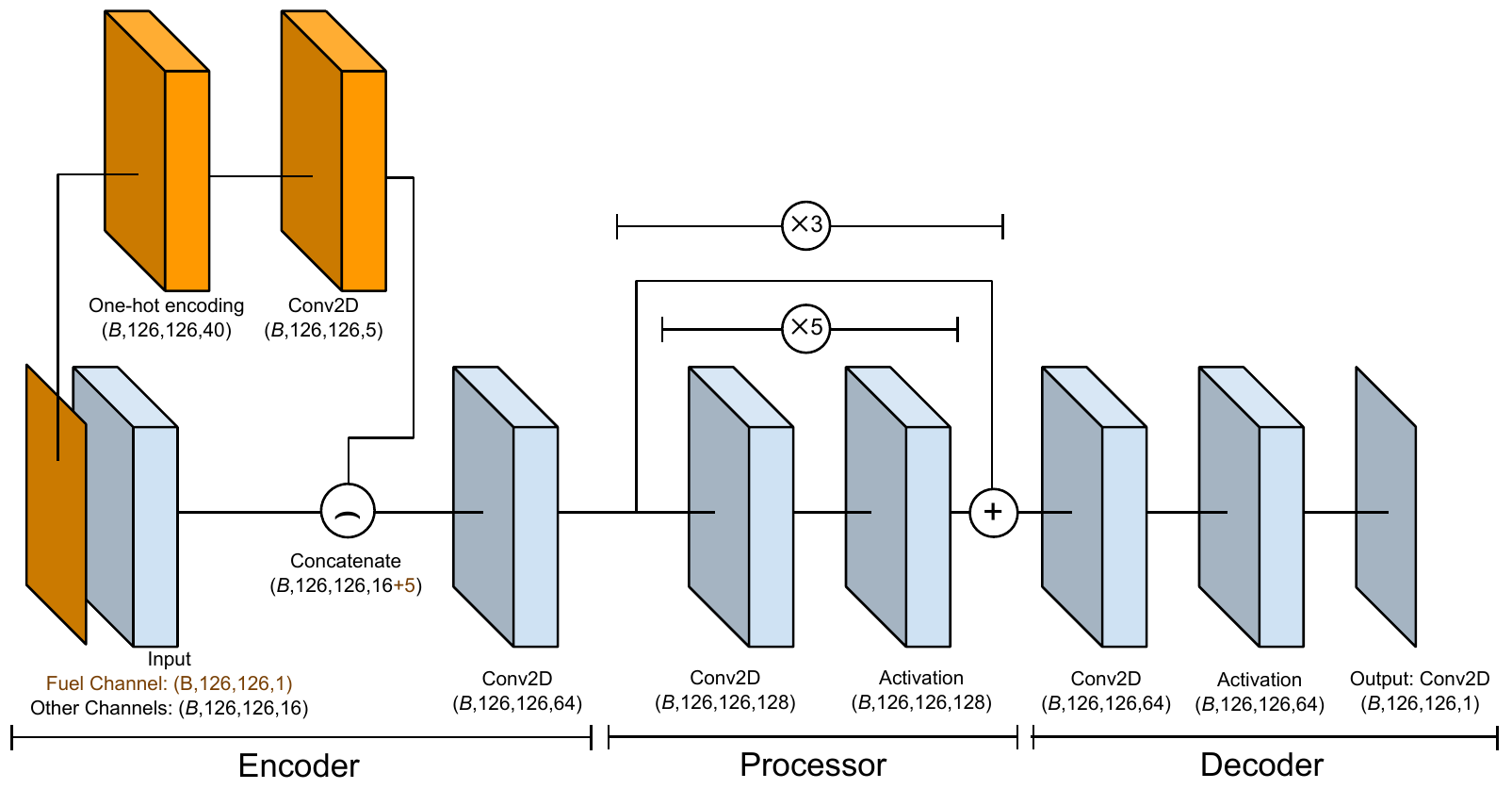}
    \caption{Schematic of EPD architecture employed in the present work. The path taken by the discretely-valued fuel channel is highlighted in orange.}
    \label{fig:epd}
\end{figure}

%%%%%%%%%%%%%%%%%%%%%
\Cref{fig:epd} provides the structure of the EPD model \cite{EPD}. The EPD's encoder first uses an embedding layer to convert the discrete fuel channel into a continuous embedding space (highlighted in orange).  Discrete channels can be challenging for DNNs to incorporate directly since small changes in the discrete value can lead to large changes in the underlying dynamics.  The embedding layer transforms the discrete fuel values into a higher dimensional latent space such that similarly behaving fuel types are close to each other in the embedding space, even if the discrete values were far apart.  This allows the information within the fuel index to be more effectively considered by subsequent layers of the DNN.

%====================
%====================

The output of the embedding layer is concatenated with the other 16 continuous input channels, and fed into a single convolutional transformation.  A total of 15 convolutional transformations are sequentially applied in the processor portion of the model.  Long sequences of transformations can be difficult to train, so skip links are added around every block of five sequential convolution layers~\cite{resnet}.  The EPD model finishes with a few convolutional transformations in the decoder.

The main differentiators between the DCIGN model and the EPD model are the removal of the fully-connected layers, the removal of the Max Pooling downsampling stages and using substantially smaller convolutional kernels.  These changes all allow the EPD model to focus more specifically on local relationships between neighboring cells.

The hyperparameters for the EPD model were empirically determined and include: the number of blocks of convolutions in the processor (3 blocks of 5 convolutions each),  the number of filters in each convolution (encoder: 64, processor: 128, decoder: 64), the activation function (encoder: identity, processor: ReLU, decoder: ReLU), the size of the convolutional kernels (encoder: 5, processor: 3, decoder: 5), the training batch size (64) and the learning rate ($10^{-4}$).

%%%%%%%%%%%%%%%%%%%%%
\subsection{\label{sec:EPD_ConvLSTM}EPD-ConvLSTM Model}
\begin{figure}[!htb!]
    \centering
    \includegraphics[width=0.95\textwidth]{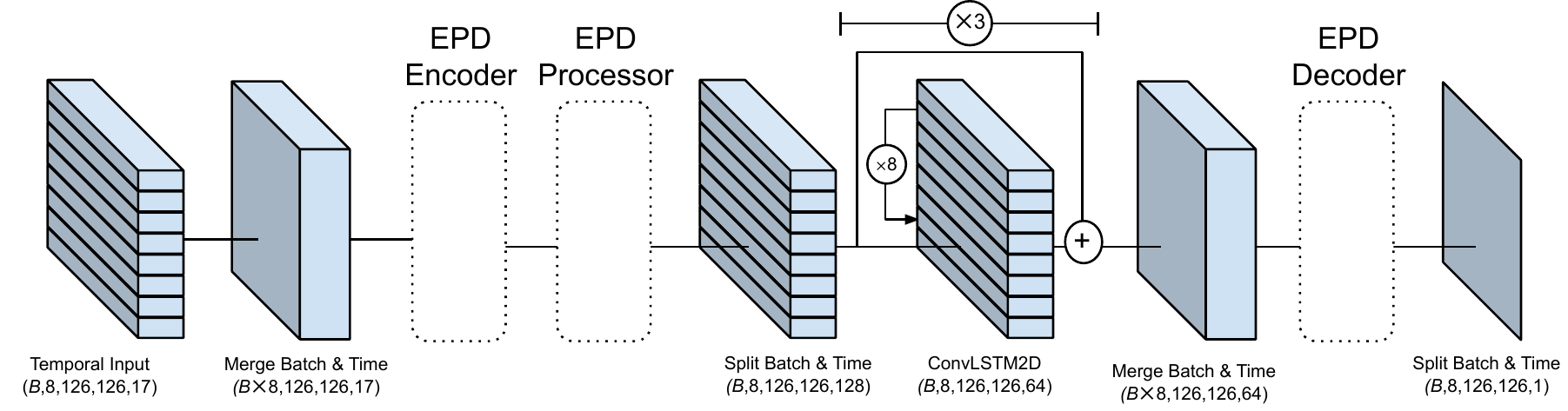}
    \caption{Schematic of EPD-ConvLSTM architecture employed in the present work. The EPD encoder, processor and decoder are the same ones provided in \cref{fig:epd}.  For steps that have a temporal dimension, this visualization breaks up the slab into 8 stacked levels to correspond to the eight consecutive time steps in the data.}
    \label{fig:epd-convlstm}
\end{figure}

%%%%%%%%%%%%%%%%%%%%%
While the EPD model effectively considers spatial relationships, it does not account for temporal relationships in the data.  For those, we use ConvLSTM recurrent transformations (see~\ref{appendix:brief dnn background}, \cite{GenerativeDeepLearning}).  By combining these approaches, we introduce the EPD-ConvLSTM model, \cref{fig:epd-convlstm}.

%====================
%====================

Compared to the EPD-model, the main difference in the EPD-ConvLSTM model is that the input contains a time series of sequential states of the fire.  But as seen in \cref{fig:epd-convlstm}, the EPD-ConvLSTM model uses the same encoder, processor and decoder as the EPD model.

The trick to getting this to work is to treat sequential steps in a time series instead as different elements within the batch dimension.  E.g., if the input to the EPD-ConvLSTM model had shape $(B{=}8, T{=}8, H{=}126, W{=}126, C{=}17)$, there are a total of 8 time series, where each time series has 8 time steps.  We can reshape this into $(B{=}64, H{=}126, W{=}126, C{=}17)$, so when the EPD layers process the data, they are simply treating time points as additional batch training points.  Any EPD stages that lead into EPD-ConvLSTM layers have their output reshaped to restore the temporal dimension.

The EPD-ConvLSTM model uses the same hyperparameters used by the EPD model, with the following additional hyperparameters that were empirically determined: the number of ConvLSTM blocks to use (3) and the number of sequential convolutions within each block (5).  Hyperparameter searches were done to determine the optimal length for the time series, but memory constraints limited the sequence to roughly 10 steps.  A length of eight was used, and by setting the batch size to 8 as well, the batches of data the EPD-ConvLSTM model trained on had the same amount of data overall as the batches of data the EPD model was trained on.

%%%%%%%%%%%%%%%%%
\section{\label{sec:Methods}Evaluation Methods}
%%%%%%%%%%%%%%%%%

In this section, we describe the evaluation process used to measure the efficacy of the DCIGN, EPD and EPD-ConvLSTM models.  We start by describing the autoregressive process used to predict the dynamics of the fire-spread behavior.  We then provide the metrics used to measure the quality of predictions, and the bootstrapping process used to determine the statistical confidences of those metrics.

%%%%%%%%%%%%%%%%%
\subsection{Autoregressive Predictions and Post-processing}
\label{sec:autoregressive_predictions_and_post_processing}
%%%%%%%%%%%%%%%%%

To assess the accuracy of our DNN models in predicting the dynamics of wildfire propagation, we measure the accuracy of each ML model over a time period of 1,500 minutes (100 autoregressive predictions).  The prediction made at time $\tau$ is used to construct the input for the prediction made at time $\tau+\Delta\tau$.  The state of the fire at time $\tau$ is denoted $X_\tau$ and the resulting prediction is denoted $Y_\tau$.  For the DCIGN and EPD models, $X_\tau$ has the shape $(B, H=126, W=126, C=17)$.  For the EPD-ConvLSTM model, $X_\tau$ has the shape $(B, T=8, H=126, W=126, C=17)$.  As all three models make the exact same prediction, the fraction of a cell's fuel that will burn away at time $\tau$, $Y_\tau$ always has the shape $(B, H=126, W=126, C=1)$.

$Y_\tau$ and $X_\tau$ are used to construct $X_{\tau+\Delta\tau}$ in the autoregressive process.  Recall from \cref{sec:Data}, that there are seventeen channels.  Fourteen of these channels are constant across time, so when constructing $X_{\tau+\Delta\tau}$, they can simply be copied from $X_\tau$.  Three of the channels represent the state of fuel: the \longvar{vegetation}, \longvar{fire_front} and \longvar{scar} channels.  We denote these channels by $X^{veg}$, $X^{front}$ and $X^{scar}$, and thus: $X_{\tau+\Delta\tau}^{front} = Y_\tau$, $X_{\tau+\Delta\tau}^{veg} = X_\tau^{veg} - Y_\tau$ and $X_{\tau+\Delta\tau}^{scar} = X_\tau^{scar} + Y_\tau$.

After each prediction, negative results were clipped to 0 to prevent invalid predictions due to round-off errors.  No other regularization or limiting operations were applied to the EPD or EPD-ConvLSTM models, though as was also observed in~\citet{Hodges2019}, the DCIGN model's predictions needed to be regularized with a mean filter using a $(3 \times 3)$ kernel.

%%%%%%%%%%%%%%%%%%%%%%%%%%%%%%%%%
\subsection{Additional Training Details}
\label{sec:training_details}
%%%%%%%%%%%%%%%%%%%%%%%%%%%%%%%%%

We following common practice and split each dataset into training, validation, and testing sets with a ratio of 80:10:10~\cite{IHME_CHUNG_MISHRA_PECS2022}, resulting in a total of 8000 unique training fire sequences, 1,000 validation fire sequences and 1,000 testing fire sequences.  Each model was trained only on data from a single dataset.  All ML models were built in {\sc TensorFlow 2.0}~\cite{tensorflow}, with {\sc{eager mode}} enabled.  Keras APIs~\cite{keras} were used to build each of the ML models.  Models were trained on a single machine with 16 CPU cores and eight NVIDIA P100 GPUs.  Training models on the \textit{single fuel} dataset took approximately 4 days for the EPD model and 6 days for EPD-ConvLSTM model.  Training on the \textit{multiple fuel} and \textit{California} datasets took 8 days for the EPD model and 13 days for the EPD-ConvLSTM model.  The {\sc Adam} optimizer \cite{adam_optimizer} was used.  For each dataset, each model was trained independently three times with the same set of hyperparameters.  Of the three runs, the run with the lowest validation loss was selected for full evaluation. Training models on the \textit{single fuel} dataset required 400 epochs to reach convergence where each epoch was trained on 1,000 batches of data.  The \textit{multiple fuel} dataset required 800 epochs.

%%%%%%%%%%%%%%%%%%%%%%%%%%%%%%%%%%%%%%
\subsection{Performance Metrics}
\label{section:performance_metrics}
%%%%%%%%%%%%%%%%%%%%%%%%%%%%%%%%%%%%%%

To quantify the accuracy of the ML model in predicting the transient fire dynamics, we consider different metrics.  Mean squared error, $MSE$, is one of the most common metrics used to measure the regression efficacy of a prediction, $Y$, for data point, $X$, with label, $\widehat{Y}$:
\begin{equation}\label{eq:ASE} 
  MSE(Y, \widehat{Y}) = \frac{1}{HW}\sum_{i=1}^{H}\sum_{j=1}^{W}(Y_{ij} - \widehat{Y}_{ij})^2\;.
\end{equation}
We use MSE as the loss function during training, but for evaluating how well a model does over an entire dataset, $D$, we use root mean squared error (RMSE).  Recall that the testing portion of each one of our three datasets contains 1,000 unique fire sequences and each of those sequences is used to construct multiple data points.  RMSE gives a general sense of the averaged error across all cells in data points in $D$:
\begin{equation}\label{eq:RMSE} 
  RMSE(D) = \frac{1}{HW|D|}\left(\sum_{Y,\widehat{Y}}^{D}\sum_{i=1}^{H}\sum_{j=1}^{W}(Y_{ij} - \widehat{Y}_{ij})^2\right)^\frac{1}{2}\;.
\end{equation}
Given that predicting the speed of the fire is often of utmost importance, we also measure the error in the size of the predicted fire front, regardless of the location, with the summed total error metric, $STE$:

\begin{equation}
\label{eq:STE_over_corpus}
  STE(D) = \frac{1}{|D|}\sum_{Y,\widehat{Y}}^D\left|\sum_{i=1}^{H}\sum_{j=1}^{W} Y_{i,j} - \sum_{i=1}^{H}\sum_{j=1}^{W} \widehat{Y}_{i,j} \right|
\end{equation}

The $STE$ metric provides a measure for the number of cells worth of fire that the prediction was off by, regardless of whether the predicted fire was in the correct location.  Thus, models that have low $STE$ are predicting the correct rate of the fire growth whereas models that have low $RMSE$ are predicting the correct location of the fire.

In addition to tracking regression statistics, we also track classification statistics by converting the regressed value into a binary classification. This is done by introducing a threshold, below which a cell is considered to be \textit{not on fire} and above which a cell is said to be \textit{on fire}.  We use a threshold of 0.1 as a general indicator that the cell at this point is quite noticeably on fire though quantitative results are robust to reasonable changes in this threshold.

We use the Jaccard Similarity Coefficient (JSC)~\cite{shape_agreement} to measure the quality of the classifications. JSC grows as the intersection between the prediction and the ground truth grows, resulting in a maximal score when the intersection and union are identical (for those that prefer the F1 score, the results were qualitatively similar).  The Jaccard score can be defined with indicator variables.  Let $I^{Y_{i,j}}$ be an indicator variable that is 1 when the value in cell $(i,j)$ is greater than the classification threshold:

\begin{align}
    \label{eq:jaccard}
    JSC(D) &= \frac{\sum_{Y,\widehat{Y}}^D\cap{(Y,\widehat{Y}})}{\sum_{Y,\widehat{Y}}^D\cup({Y,\widehat{Y}})}\;\\
    \cap{(Y,\widehat{Y}}) &= \sum_{i=1}^{H}\sum_{j=1}^{W}I^{Y_{ij}} \wedge I^{\widehat{Y}_{ij}}\;\\
    \cup{(Y,\widehat{Y}}) &= \sum_{i=1}^{H}\sum_{j=1}^{W}I^{Y_{ij}} \vee I^{\widehat{Y}_{ij}}\;.
\end{align}

JSC ranges between 0.0, in which the prediction and the ground truth do not overlap, and 1.0, in which the ground truth and prediction are perfectly aligned.

In addition to tracking the location of the fire \textit{front} over time, the location of the resulting \textit{scar} left by the fire is also tracked. All of the same statistics are evaluated for the fire scar.

%%%%%%%%%%%%%%%%%%%%%%%%%%%%%%%%%%%%%
\subsection{Bootstrapping Confidence Intervals}
\label{sec:boot}
%%%%%%%%%%%%%%%%%%%%%%%%%%%%%%%%%%%%%%

When assessing the performance of the different ML models on a given metric, it is important to determine whether an observed difference is real or coincidentally caused by the stochastic process of placing data in the testing, training and validation datasets.  To account for this, we use bootstrapping \cite{tibshirani_1993} to estimate confidence intervals for each metric.   For each of the three test datasets, (\textit{single fuel}, \textit{multiple fuel}, \textit{California}), we build 20 new resampled datasets.  Each resampled dataset is constructed by randomly sampling fire sequences from the 1,000 sequences in the original testing dataset, with replacement, until the bootstrapping dataset also contains 1,000 fire sequences.  Data points are then generated from the fire sequences in the resampled dataset, and the metric is computed over those data points.  After this is repeated 20 times, there is a distribution of values for the metric.  We compute 50\% confidence intervals by taking the 0.25 and 0.75 quantiles of the distribution. 90\% confidence intervals are computed by taking the 0.05 and 0.95 quantiles.

%%%%%%%%%%%%%%%%%%%%%%%%%%%%%%%%%%%%%%
\section{Results}
\label{sec:Results}
%%%%%%%%%%%%%%%%%%%%%%%%%%%%%%%%%%%%%%
\subsection{Time-to-arrival Maps}
\label{sec:tta}
%%%%%%%%%%%%%%%%%%%%%%%%%%%%%%%%%%%%%%

In this section, we present results of autoregressive predictions from the DNN models via time-to-arrival (TTA) maps.  These maps demonstrate how the models have learned to realistically represent the complex physical interaction between the fire front and aspects of the environment such as fuel type, terrain-slope and wind.  While they are useful for visualizing a sequence of predictions, note that TTA maps do not capture the full dynamics of the simulation.  Once a cell starts burning, the dynamics of how that burning occurs are considered by the DNN models, but opaque to TTA maps.

%%%%%%%%%%%%%%%%%%%%%%%%%%%%%%%%%%%%%%
\subsubsection{DCIGN Model}
%%%%%%%%%%%%%%%%%%%%%%%%%%%%%%%%%%%%%%

\begin{figure}[!htb!]
    \centering
    \includegraphics[scale=0.45]{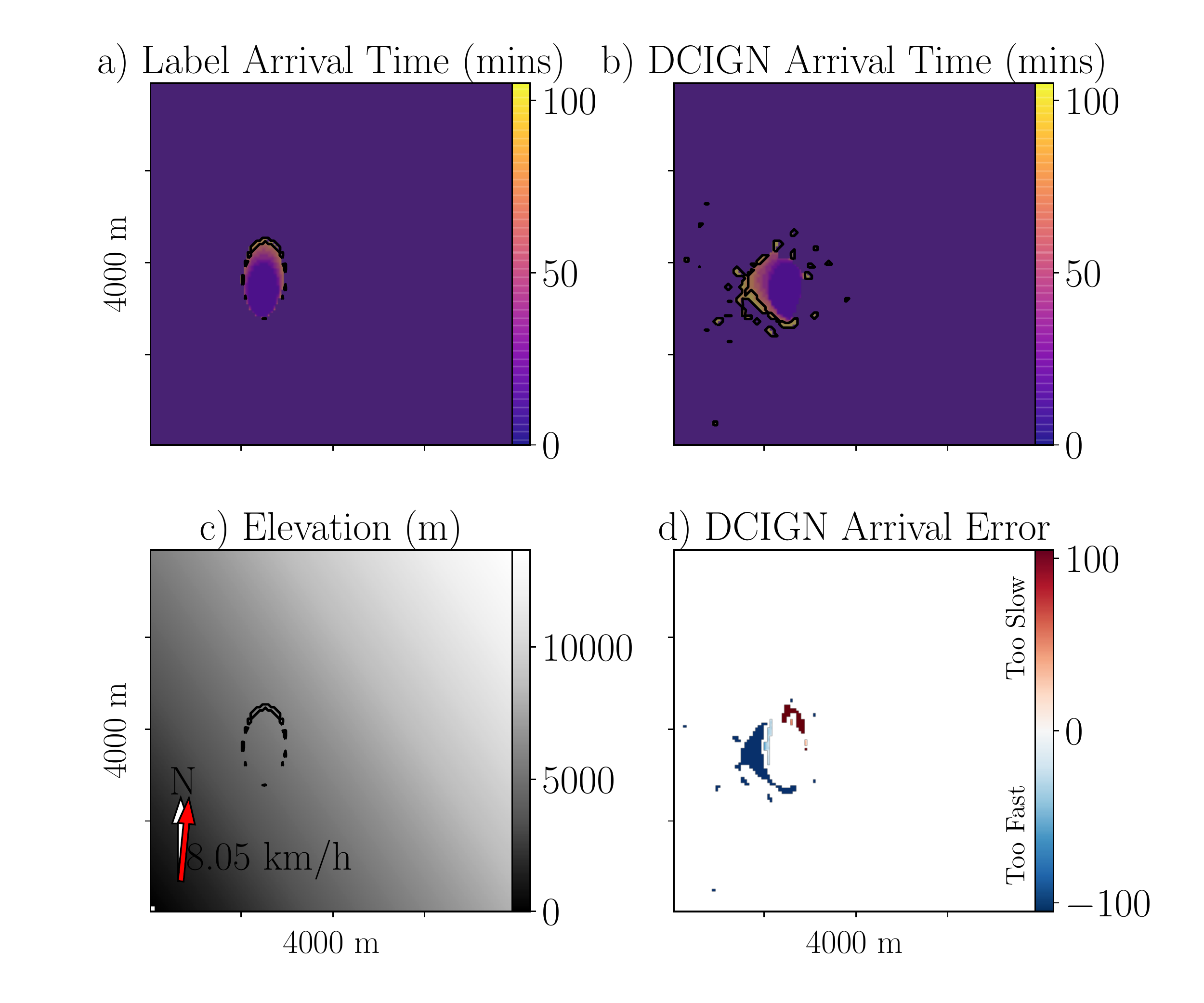}
    \caption{TTA maps for DCIGN model on a randomly selected fire sequence in the \textit{single fuel} test dataset. a) Arrival time in the ground truth generated by FARSITE; background color denotes that the homogeneous fuel type is GR1. b) Arrival time predicted by the DCIGN model. c)  Elevation and wind with TTA contours. d) Error in arrival time prediction.  Blue and red colors denote overprediction and underprediction of fire spreading, respectively.}
    \label{fig:dcign_tta}
\end{figure}

\Cref{fig:dcign_tta} shows TTA maps for a typical fire sequence in the \textit{single fuel} dataset generated by the DCIGN model.  The DCIGN model immediately made errors in predicting the direction the wildfire propagates.  After just 120 minutes (eight autoregressive predictions), the DCIGN model started making unrealistic predictions of spurious fire spotting in regions far away from the actual fire front.  The contour lines in \cref{fig:dcign_tta}a) show the small distance the fire travelled in the FARSITE simulation, whereas the contour lines in \cref{fig:dcign_tta}b) highlight the errors the DCIGN model made during that same time frame.

%=====================
%=====================

Qualitatively, results were similar on the \textit{multiple fuel} and \textit{California} datasets and while our work reveals some shortcomings of the DCIGN model in predicting short-time fire dynamics, it is important to note that it was successfully employed for predictions at larger time scales (6 or 24 hours)~\cite{Hodges2019,firecast}.

%%%%%%%%%%%%%%%%%%%%%%%%%%%%%%%%%%%%%%
\subsubsection{EPD and EPD-ConvLSTM Models}
%%%%%%%%%%%%%%%%%%%%%%%%%%%%%%%%%%%%%%

\begin{figure}[htb!]
    \centering
    \includegraphics[width=\textwidth]{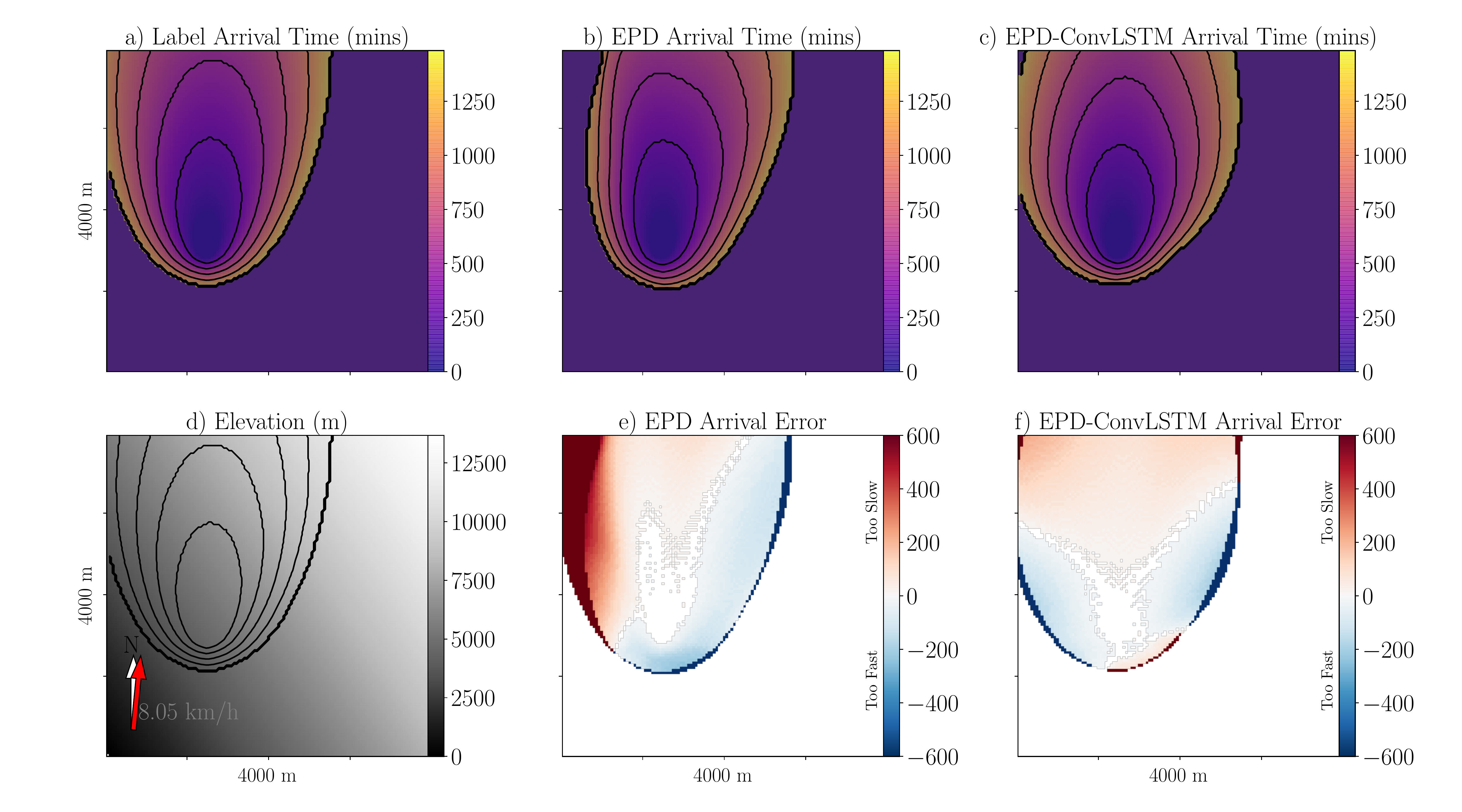}
    \caption{TTA maps for EPD and EPD-ConvLSTM models on the same \textit{single fuel} fire sequence seen in \cref{fig:dcign_tta} after 1,500 minutes.  a) TTA map for the ground truth generated by FARSITE; background color denotes that the homogeneous fuel type is GR1. TTA maps predicted by b) EPD model and c) EPD-ConvLSTM model. d) Elevation of the field and direction of the wind with the TTA contours from a). Errors in e) the EPD prediction and f) the EPD-ConvLSTM prediction.}
    \label{fig:epd_epdc_constant_field_tta}
\end{figure}

\Cref{fig:epd_epdc_constant_field_tta} provides results for the EPD and the EPD-ConvLSTM models on predicting the same \textit{single fuel} fire sequence seen in \cref{fig:dcign_tta}, but after 1,500 minutes (100 autoregressive predictions) instead of just 90 minutes.  Unlike the DCIGN model, both the EPD and EPD-ConvLSTM models made realistic, accurate and stable predictions.  Although there are no physical constraints enforced in the training of these models, both EPD and EPD-ConvLSTM models were able to reproduce the elliptical fire scars that are expected in field fires with constant wind and planar sloped terrain. In this context, we note that the introduction of physical principles during the training is expected to further improve the model accuracy and reduces the amount of data needed for training~\cite{IHME_CHUNG_MISHRA_PECS2022}.

%=====================
%=====================

\begin{figure}[htb!]
    \centering
    \includegraphics[width=\textwidth]{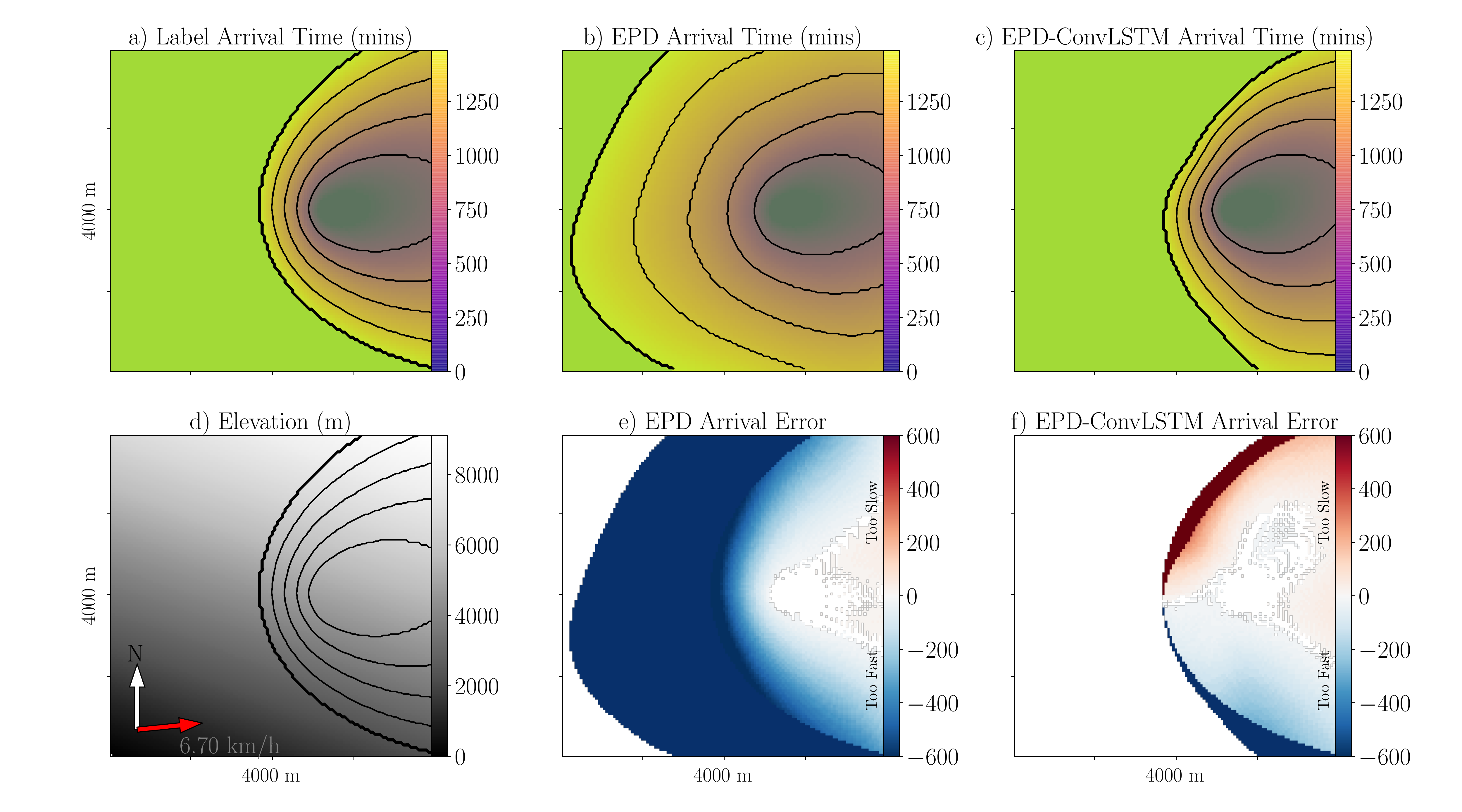}
    \caption{TTA maps for EPD and EPD-ConvLSTM models on a fire sequence in the \textit{multiple fuel} dataset. Panels follow the same layout as in \cref{fig:epd_epdc_constant_field_tta}.}
    \label{fig:epd_epdc_dynamic_field_tta}
\end{figure}

\Cref{fig:epd_epdc_dynamic_field_tta} provides results for the EPD and EPD-ConvLSTM models in predicting the dynamics of a fire sequence in the \textit{multiple fuel} dataset. Because there are variations in the fuel types used across fire sequences in this dataset, training and prediction are more difficult than in the \textit{single fuel} dataset.  In~\cref{fig:epd_epdc_dynamic_field_tta}, both the EPD and EPD-ConvLSTM models make some errors in their predictions. The EPD model struggles to maintain the proper elliptical shape of the fire front, and overpredicts the fire-spread rate.  Conversely, predictions from the EPD-ConvLSTM model achieves a more elliptical shape and more accurately estimates the fire-spread rate.  As expected, errors are more significant than in the \textit{single fuel} dataset.

%=====================
%=====================

%=====================
\begin{figure}[htb!]
    \centering
    \includegraphics[width=\textwidth]{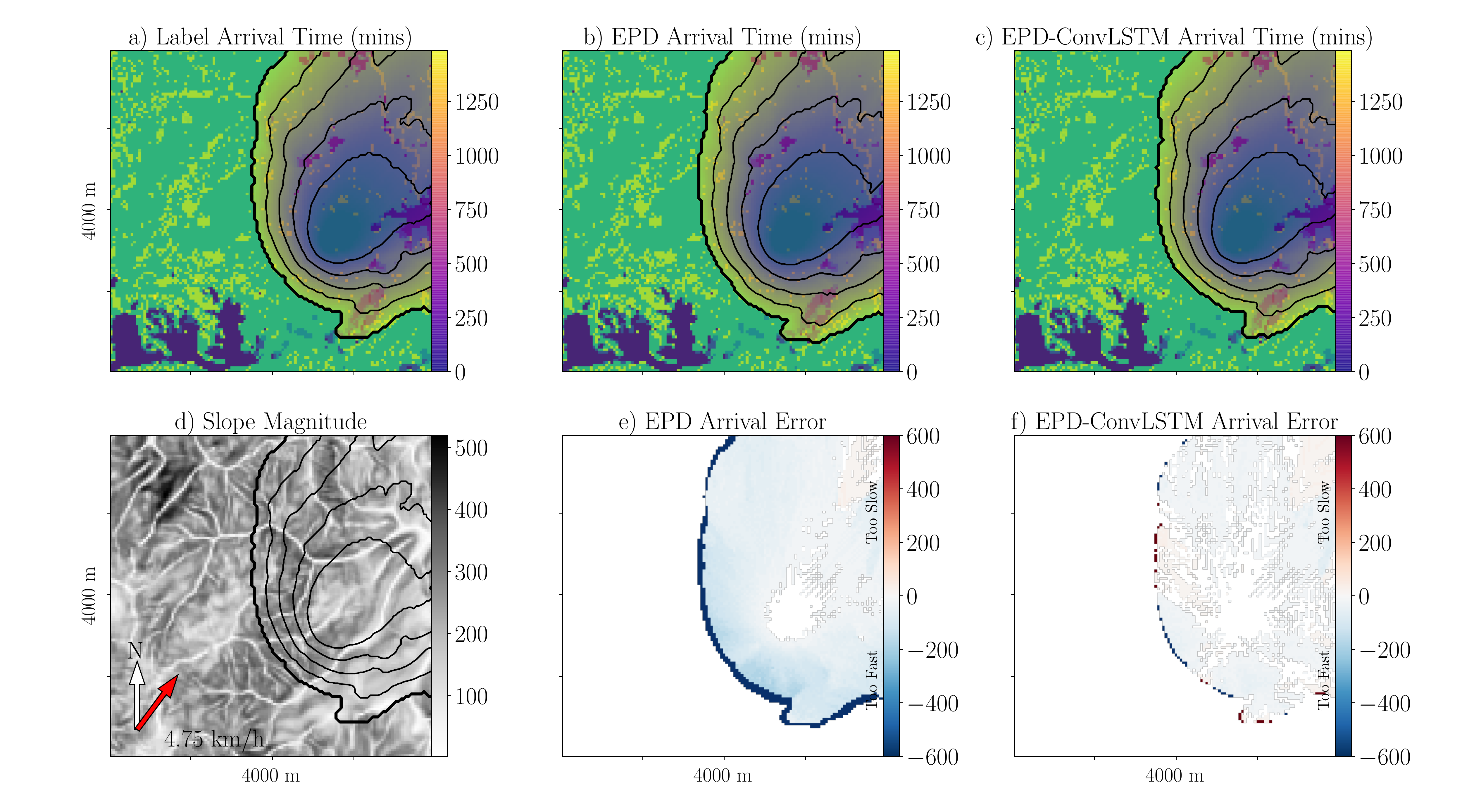}
    \caption{TTA maps for the EPD and EPD-ConvLSTM model on a fire sequence in the \textit{California} dataset.  Panels follow the same layout as in \cref{fig:epd_epdc_constant_field_tta}.}
    \label{fig:epd_epdc_realistic_85}
\end{figure}
%=====================

\Cref{fig:epd_epdc_realistic_85} shows a comparison between the EPD and EPD-ConvLSTM models' predictions on a fire sequence randomly selected from the \textit{California} dataset. The majority of the field is covered by timber litter fuel (green), with many small patches of slash burn fuel (light green).  Thus, the general shape of the fire scar is roughly elliptical.  The wind pushes the fire in the northeasterly direction but there are patches of grass shrub (light red) in the southern and northern regions where the fire accelerates more quickly than on the timber litter fuel.  The southern patch of grass shrub is located on terrain that slopes upward, further accelerating the fire in the opposite direction of the wind.  This acceleration can clearly be seen in the TTA contour lines near the bottom of the field.

Both the EPD and EPD-ConvLSTM models performed well and correctly considered the complex interactions between the various fuel types, the changes in elevation and the wind.  The EPD model did not make any egregious errors, but did not estimate the speed of the fire propagation as accurately as the EPD-ConvLSTM model.  The dark blue boundary in \cref{fig:epd_epdc_realistic_85}e) denotes cells the EPD model predicted the fire would reach after 1,500 minutes, but never actually did in the ground truth (false positives).  Conversely, in \cref{fig:epd_epdc_realistic_85}f), there are far fewer dark blue cells, but there are also some dark red cells,  indicating locations the fire was supposed to reach, but were never predicted to do so (false negatives).

%=====================
\begin{figure}[htb!]
    \centering
    \includegraphics[width=\textwidth]{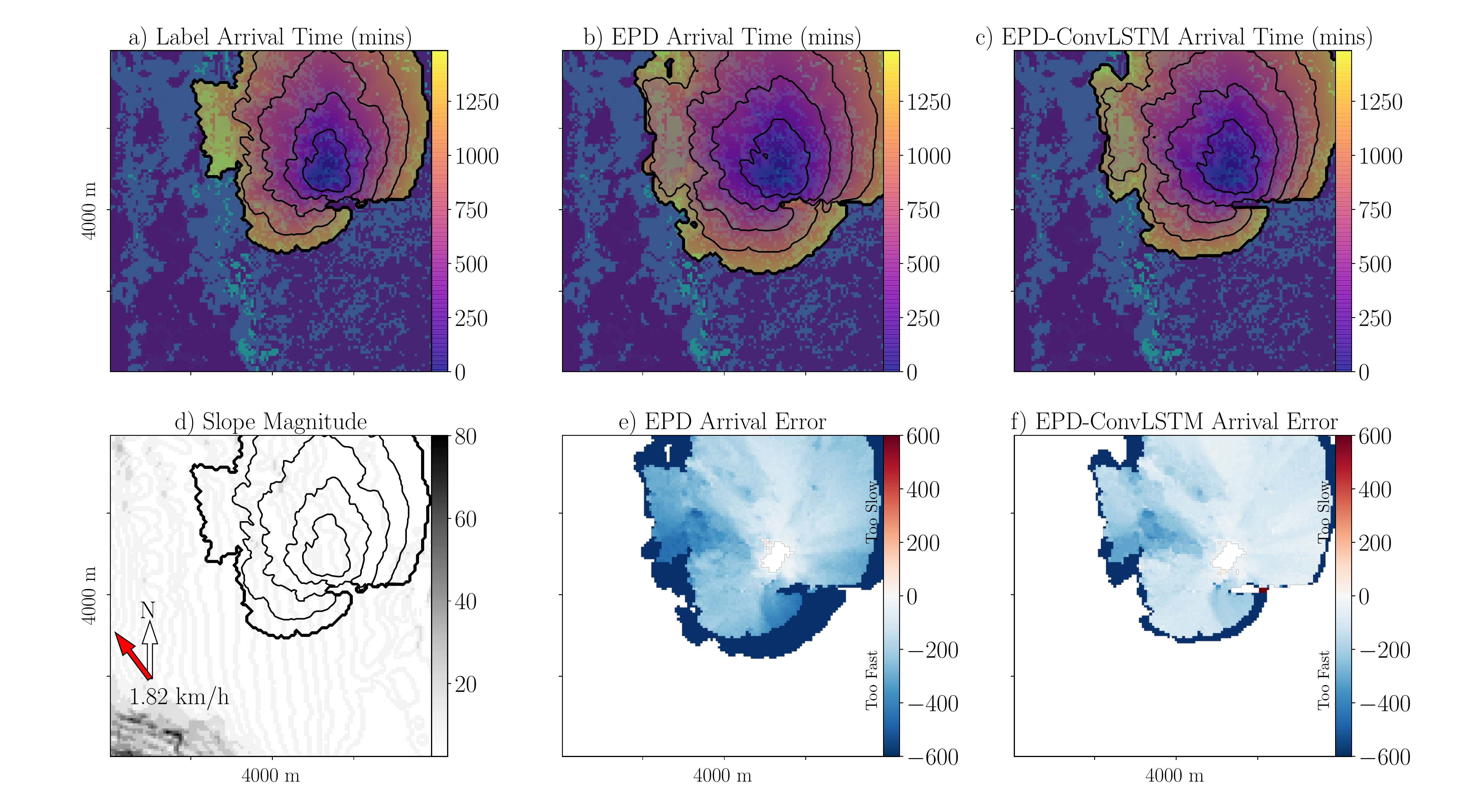}
    \caption{TTA maps for the EPD and EPD-ConvLSTM model on a second fire sequence in the \textit{California} dataset that contained a patch of nonburnable fuel. Panels follow the same layout as in \cref{fig:epd_epdc_constant_field_tta}}
    \label{fig:epd_epdc_realistic_100}
\end{figure}
%=====================

\Cref{fig:epd_epdc_realistic_100} shows predictions from the EPD and EPD-ConvLSTM models for a second fire sequence in the \textit{California} dataset.  The field is primarily a combination of short grass fuels (dark blue) and short shrubs (light blue).  There are also small patches of dense shrubs (light green) arranged in a thin strip spanning the entire height of the field.  There is a large patch of these dense shrubs towards the top of the field and the contour lines in \cref{fig:epd_epdc_realistic_100}a) clearly show a dramatic increase in the speed of the fire once it hits this dense patch of vegetation.

The field also contains a small strip of nonburnable open land on the right side of the field.  This open land is consistent with a road that dead-ends into a few buildings, though the exact nature of the nonburnable land is not fully specified.  The individual cells in the field that correspond to road cannot be readily made out in the figure, but the effects they have on the TTA map are immediately apparent.  Once the fire reaches the road, the fire completely ceases to further propagate.  Eventually, the fire does wrap around the dead-end in the road and begins engulfing the remaining fuel to the south.

As in the previous \textit{California} example, both the EPD and EPD-ConvLSTM models did well in predicting the behavior of the fire front and location of the final fire scar.  Once again, the EPD-ConvLSTM model had lower overall error, though both models did overestimate the speed of the fire front.  The EPD model did end up errantly burning the road, but only after the fire had wrapped around the road and begun burning fuel on both sides of the road.  The EPD-ConvLSTM model correctly avoided burning the road.

\subsection{Full Evaluation}
\label{sec:full_evaluation}

The individual TTA map examples provided in \cref{sec:tta} were selected to provide a visual illustration of how well each model does on average across each of the three datasets after 1,500 minutes.  Since the examples depict the final fire scars, we can contrast the JSC for the example with the average JSC across the entire test dataset.  For instance, the JSC for the EPD model's prediction in \cref{fig:epd_epdc_constant_field_tta} is 0.86, whereas on average across the entire \textit{single fuel} dataset, it is 0.82.  Across all TTA maps shared in  \cref{fig:epd_epdc_constant_field_tta,fig:epd_epdc_dynamic_field_tta,fig:epd_epdc_realistic_85}, the final fire scar's JSC is in within 0.04 of the average JSC across the entire test dataset, with two exceptions.  First, the EPD results in \cref{fig:epd_epdc_dynamic_field_tta} shows a particularly poor prediction (0.47 JSC versus 0.74 on average) and \cref{fig:epd_epdc_realistic_85} shows a particularly good prediction (0.99 JSC versus 0.94 on average).

%\subsubsection{Evaluation}

\begin{figure}[htb!]
    \centering
    \begin{minipage}{0.95\textwidth}
      \includegraphics[width=\textwidth]{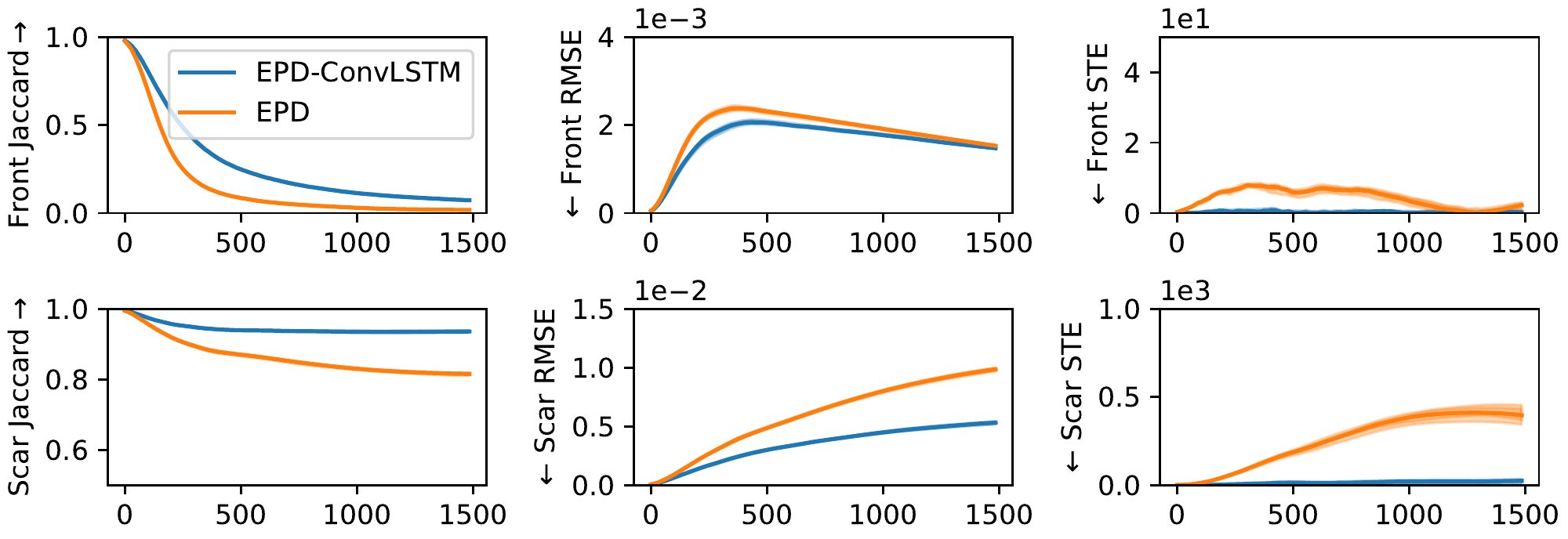}
      \caption*{a) \textit{single fuel} dataset.}
    \end{minipage}
    \begin{minipage}{0.95\textwidth}
      \includegraphics[width=\textwidth]{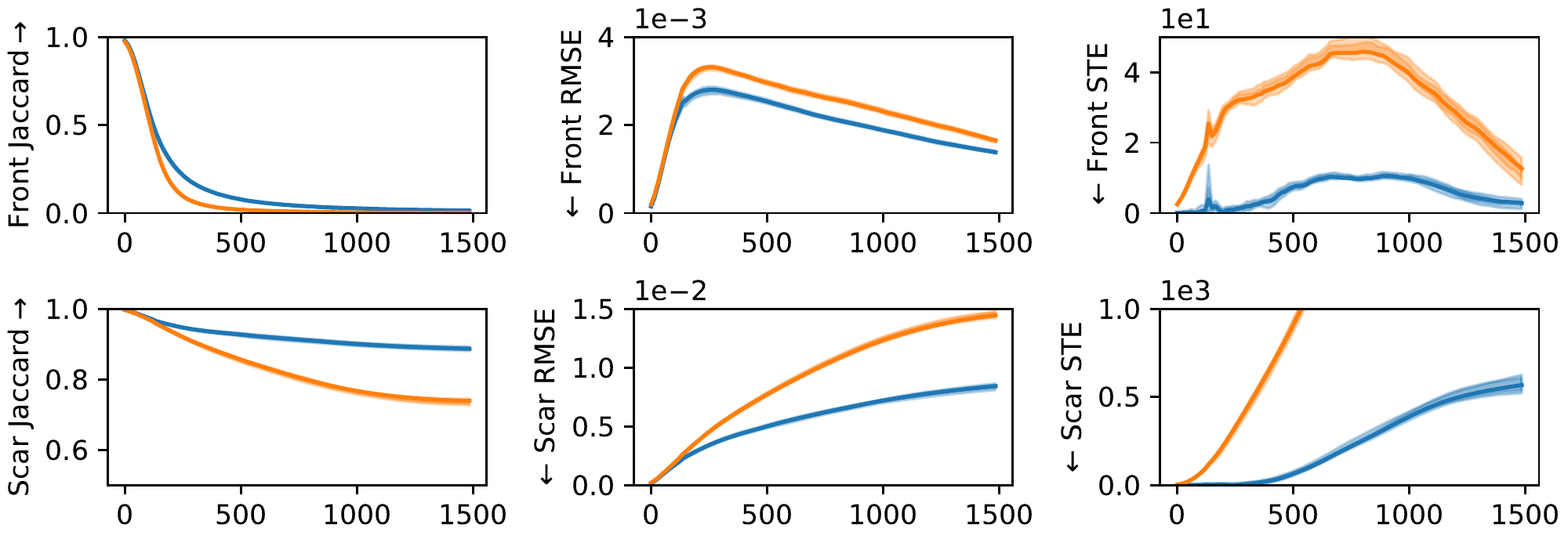}
      \caption*{b) \textit{multiple fuel} dataset.}
    \end{minipage}
    \begin{minipage}{0.95\textwidth}
      \includegraphics[width=\textwidth]{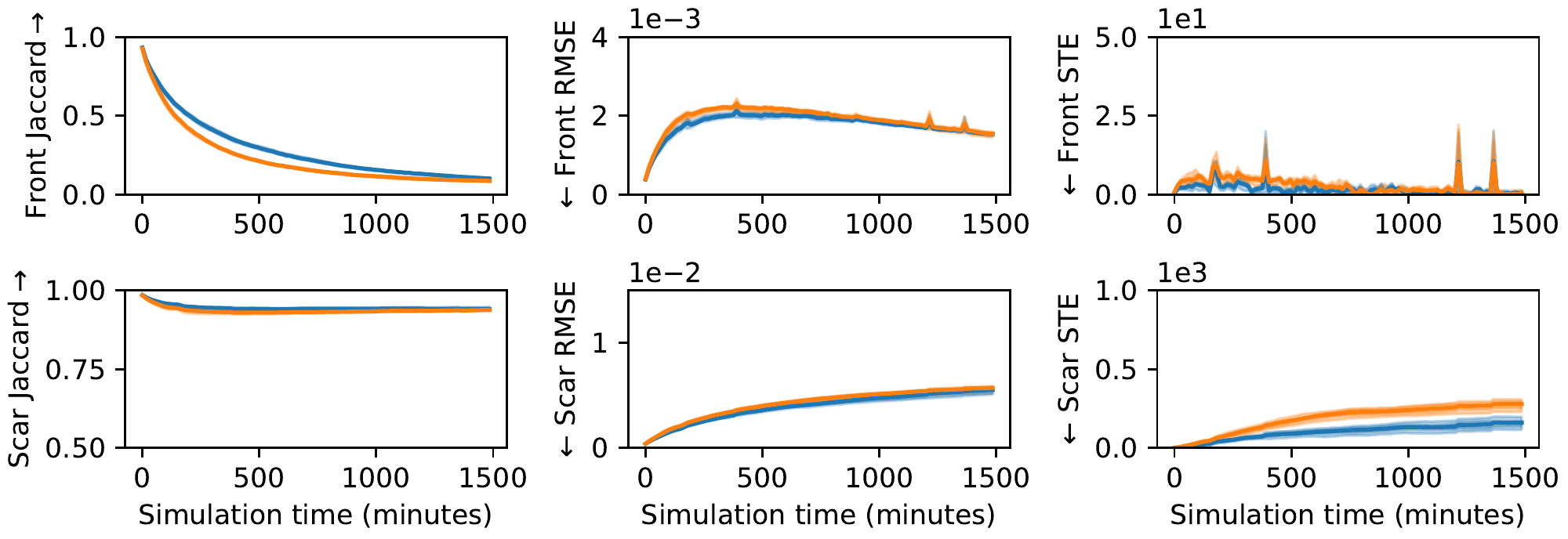}
      \caption*{c) \textit{California} dataset.}
    \end{minipage}
    \caption{Full evaluation of the EPD-ConvLSTM and EPD models.}
    \label{fig:full_eval_graphs}
\end{figure}

\begin{table}[htb!]
\footnotesize
\begin{center}
\begin{sideways}
\begin{tabular}{|c|c|c|c|c|c|c|c|c|c|c|c|c|}
\hline
 & \multicolumn{6}{c|}{EPD-ConvLSTM} & \multicolumn{6}{c|}{EPD} \\
\hline\hline
 & Front & Scar & Front & Scar & Front & Scar & Front & Scar & Front & Scar & Front & Scar \\
Step & Jcd & Jcd & STE & STE & RMSE & RMSE & Jcd & Jcd & STE & STE & RMSE & RMSE \\
\hline
\multicolumn{13}{|c|}{\textit{single fuel}} \\
\hline
1   & 0.98 & 1.00 & 1.9E--2 & 1.9E--2 & 4.7E--5 & 4.7E--5 & 0.98 & 1.00 & 3.2E--1 & 3.2E--1 & 5.7E--5 & 5.7E--5  \\
25  & 0.33 & 0.94 & 6.4E--1 & 8.9e+0 & 2.0E--3 & 2.4E--3 & 0.13 & 0.88 & 7.6E+0 & 1.3E+2 & 2.4E--3 & 3.9E--3 \\
50  & 0.16 & 0.94 & 2.6E--1 & 1.3E+1 & 1.9E--3 & 3.8E--3 & 0.05 & 0.85 & 6.4E+0 & 2.9E+2 & 2.1E--3 & 6.6E--3 \\
75  & 0.10 & 0.93 & 9.0E--2 & 2.0E+1 & 1.7E--3 & 4.8E--3 & 0.02 & 0.82 & 1.4E+0 & 4.0E+2 & 1.8E--3 & 8.6E--3 \\
100 & 0.07 & 0.94 & 2.5E--1 & 2.4E+1 & 1.5E--3 & 5.3E--3 & 0.02 & 0.82 & 2.2E+0 & 3.9E+2 & 1.5E--3 & 9.9E--3 \\
\hline
\multicolumn{13}{|c|}{\textit{multiple fuel}} \\
\hline
1   & 0.98 & 1.00 & 4.6E--2 & 4.6E--2 & 1.5E--4 & 1.5E--4 & 0.97 & 1.00 & 2.5E+0 & 2.5E+0 & 2.0E--4 & 2.0E--4 \\
25  & 0.12 & 0.94 & 3.2E+0 & 1.9E+1 & 2.7E--3 & 4.3E--3 & 0.04 & 0.89 & 3.4E+1 & 6.5E+2 & 3.2E--3 & 6.2E--3 \\
50  & 0.04 & 0.91 & 9.9E+0 & 2.2E+2 & 2.2E--3 & 6.2E--3 & 0.01 & 0.81 & 4.5E+1 & 1.6E+3 & 2.6E--3 & 1.0E--2 \\
75  & 0.02 & 0.90 & 7.6E+0 & 4.6E+2 & 1.7E--3 & 7.6E--3 & 0.00 & 0.75 & 3.3E+1 & 2.6E+3 & 2.1E--3 & 2173--2 \\
100 & 0.01 & 0.89 & 2.8E+0 & 5.6E+2 & 1.4E--3 & 8.4E--3 & 0.00 & 0.74 & 1.2E+1 & 3.1E+3 & 1.7E--3 & 1.4E--2 \\
\hline
\multicolumn{13}{|c|}{\textit{California}} \\
\hline
1   & 0.93 & 0.99        & \tiny{4.3E--1} & 4.3E--1 & 3.8E--4 & 3.8E--4 & 0.93 & 0.98        & \tiny{8.2E--1}    & 8.2E--1 & 3.9E--4 & 3.9E--4 \\
25  & 0.36 & 0.94        & \tiny{2.0E+0} & 7.1E+1 & 2.0E--3 & 3.1E--3 & 0.27 & 0.93        & \tiny{4.3E+0}    & 1.3E+2 & 2.2E--3 & 3.5E--3 \\
50  & 0.21 & 0.94        & \tiny{1.5E+0} & 1.1E+2 & 1.9E--3 & 4.2E--3 & 0.15 & 0.93        & \tiny{2.2E+0}    & 2.2E+2 & 2.1E--3 & 4.7E--3 \\
75  & 0.14 & \tiny{0.94} & \tiny{3.7E--1} & 1.3E+2 & 1.7E--3 & 5.0E--3 & 0.10 & \tiny{0.94} & \tiny{1.4E+0}    & 2.4E+2 & 1.8E--3 & 5.3E--3 \\
100 & 0.10 & \tiny{0.94} & \tiny{4.1E--2} & 1.5E+2 & 1.5E--3 & 5.5E--3 & 0.09 & \tiny{0.94} & 6.2E--1           & 2.7E+2 & 1.5E--3 & 5.7E--3 \\
\hline
\end{tabular}
\end{sideways}
\end{center}
\caption{Full evaluation of the EPD-ConvLSTM and EPD models on all three datasets.  Small fonts indicates values with overlapping confidence intervals.  In all other results, the EPD-ConvLSTM statistically significantly outperformed the EPD model.}
\label{table:stats}
\end{table}

\Cref{fig:full_eval_graphs} provides the full evaluation for the EPD-ConvLSTM and EPD models on all three test datasets.  Solid lines correspond to results computed from one of the test datasets, whereas the dark-shaded and light-shaded regions correspond to 50\% and 90\% confidence intervals (respectively) estimated from the 20 bootstrapping datasets described in \cref{sec:boot}.  Each column of graphs provides the results across all datasets for a single metric.  The top row of graphs in each section provides results for the fire \textit{front} predictions, and the bottom row of graphs provides results for the fire \textit{scar} predictions.  The same graph in each of the three sections uses the same scale for the vertical axis, allowing the same metric to easily be compared across the three datasets. \cref{table:stats} provides the numeric values for each metric on the 1st, 25th, 50th, 75th and 100th autoregressive predictions.

The fire front JSC and fire scar JSC values (first column) are the classification-based metrics that we used to evaluate the models.  The results across all three datasets demonstrated that the EPD-ConvLSTM model consistently outperformed the EPD model.  Across all six graphs, the EPD-ConvLSTM's JSC was statistically significantly higher than the corresponding EPD model's score, with the exception of the 75th and 100th JSC on the fire front (the lack of statistical significance here is denoted by the smaller font in \cref{table:stats}).  This indicates that the EPD-ConvLSTM model did a better job selecting the specific cells that the fire-front spread transitioned to during each of the autoregressive predictions.  Also note the difference between the overall JSC between the fire front and the fire scar across all three datasets.  The fire front is a more difficult prediction to make, as a fire front might only have a width of several cells on a field with over 100 cells in both directions.  The fire scar remains much larger.  Further, once a cell is in the fire scar, it stays there, making the dynamics easier to predict.

The second column of graphs provides the RMSE metric.  As expected, the RMSE results are qualitatively similar to the Jaccard results.  The pattern of increased performance on this metric roughly mirrors the same increased performance seen in the JSC results.  However, the RMSE is more sensitive than the JSC in that it considers how far off an errant cell's prediction is, not just whether the prediction is correct versus incorrect.  This extra sensitivity resulted in the RMSE score being statistically significantly better for the EPD-ConvLSTM model across all time points in the autoregressive process.

The third column of graphs provides the results for the STE metric.  This metric measures the overall ability for the ML models to predict the correct size of the fire front and fire scar, regardless of whether the location was correct.  On predicting the overall size of the fire scar, the EPD-ConvLSTM model clearly outperformed the EPD model on all three datasets.  On predicting the overall size of the fire front, the EPD-ConvLSTM model clearly outperformed the EPD model on the \textit{single fuel} and \textit{multiple fuel} datasets, but only outperformed the EPD model in the first 50 autoregressive predictions on the \textit{California} dataset.  The strong performance of the EPD-ConvLSTM model on this metric indicates that adding temporal layers to the EPD model results in the model more successfully estimating the speed at which the fire propagates.

We had anticipated that the \textit{California} dataset would be the most difficult to predict.  It contains realistic and complex arrangements of vegetation, moisture content, elevation, etc., whereas the other two datasets contain entirely homogeneous distributions of fuel on planar landscapes.  However, the realistic landscapes result in patches of terrain that often significantly constrain the direction and speed that a fire can reasonable propagate.  It was not hard for either model to learn, e.g., that fires propagate more slowly through some fuels than others, and patches with either slow-burning fuel or even nonburnable fuel are easier to predict than the freedom the fire has to move in the \textit{single fuel} and especially \textit{multiple fuel} fires.  

Indeed, the most difficult dataset to classify was the \textit{multiple fuel} dataset.  Consider that if a particular cell is on fire in a \textit{multiple fuel} fire, it will propagate to some degree into all eight neighboring cells.  Our models are, intentionally, entirely empirical and make no assumptions about the shape or propagation dynamics that the fire will take.  This allows the same method for training a model to be learned on any of the fire datasets. 

While the accuracy of the EPD model is clearly improved with the addition of the temporal ConvLSTM layers, that improvement comes at the cost of increased inference time.  The EPD model alone takes approximately 7ms to perform a single inference.  The EPD-ConvLSTM model takes approximately 61ms to perform the same inference.  FARSITE takes anywhere between 20\,ms and 100\,ms (depending on how much fire is in the patch).

\section{Conclusion}
\label{sec:Conclusion}

Our primary contribution is evaluating the efficacy of recurrent convolutional neural networks to predicting the time-dependent behavior of wildfires. Previous studies demonstrated that convolutional neural networks can make effective predictions of the fire scar in one single large time step \cite{Hodges2019,firecast}.  We instead focus on an autoregressive process in which many small predictions are sequentially made to predict the fire scar after a large amount of time.  This method allows for the model to continuously adapt to changing dynamic environment conditions and more precisely models the evolution of the fire front by not only modeling when the fire reached a particular location, but also by explicitly predicting how much fuel burns at each location at each time step.

We demonstrate that a popular previously employed model for similar predictions, the Deep Convolutional Inverse Graphics Network (DCIGN) model, was not well-suited for making the fine-grained temporal autoregressive predictions, as the transformations in that model facilitate large-scale changes to the input state, resulting in the model generalizing poorly in this context.  We replace the transformations in the DCIGN model with more suitable transformations, leading basically to the same structure seen in the Encoder-Processor-Decoder (EPD) model.  We demonstrate that the EPD model is stable and can realistically propagate a fire front forward in time for upwards of 100 autoregressive predictions, achieving Jaccard scores of 0.82, 0.74 and 0.94 on our three datasets.  We also added recurrent layers to the EPD model, resulting in the EPD-ConvLSTM model, which significantly improves the accuracy of the predictions.

There are two primary directions we see for future work.  The first is to continue exploring the broad space of DNNs to wildland fire propagation predictions~\cite{IHME_CHUNG_MISHRA_PECS2022}.  Directions we feel are promising include 1) investigating alternative approaches to folding the autoregressive process into the loss function (potentially without recurrent layers added to the network), 2) introducing physical constraints, 3) quantifying uncertainties of ML-model predictions, and 4) exploring latent diffusion models, which have recently shown great promise in domains with complex spatial structure \cite{stable_diffusion}.

The second direction involves extending this work to more practical real-world scenarios in which observational data is far more sparse than the data generated by FARSITE.  It is unrealistic to assume that real-world data can be rich enough to train models like the EPD or EPD-ConvLSTM model. Physics-informed deep learning approaches and the consideration of data from high-fidelity physics simulations can potentially close this gap by augmenting sparse observational data. 

\section*{Acknowledgements}

Upon approval of publication, we will make all three datasets (including training, validation and testing sections), and each trained model, available for download on Kaggle.
% There are no funding sources to acknowledge. 

\bibliographystyle{elsarticle-num-PROCI}
\bibliography{firetech}

\begin{thebibliography}{37}
\expandafter\ifx\csname natexlab\endcsname\relax\def\natexlab#1{#1}\fi
\providecommand{\bibinfo}[2]{#2}

\bibitem[{Covington and Moore(1994)}]{covington_southwestern_1994}
\bibinfo{author}{W.~W. Covington}, \bibinfo{author}{M.~M. Moore},
  \bibinfo{title}{Southwestern ponderosa forest structure: {Changes} since
  {Euro}-{American} settlement}, \bibinfo{journal}{Journal of Forestry}
  \bibinfo{volume}{92} (\bibinfo{year}{1994}) \bibinfo{pages}{39--47}.

\bibitem[{Westerling et~al.(2006)Westerling, Hidalgo, Cayan, and
  Swetnam}]{westerling_warming_2006}
\bibinfo{author}{A.~L. Westerling}, \bibinfo{author}{H.~G. Hidalgo},
  \bibinfo{author}{D.~R. Cayan}, \bibinfo{author}{T.~W. Swetnam},
  \bibinfo{title}{Warming and earlier spring increase {Western} {U.S}. forest
  wildfire activity}, \bibinfo{journal}{Science} \bibinfo{volume}{313}
  (\bibinfo{year}{2006}) \bibinfo{pages}{940--943}.

\bibitem[{Mueller et~al.(2020)Mueller, Thode, Margolis, Yocom, Young, and
  Iniguez}]{mueller_climate_2020}
\bibinfo{author}{S.~E. Mueller}, \bibinfo{author}{A.~E. Thode},
  \bibinfo{author}{E.~Q. Margolis}, \bibinfo{author}{L.~L. Yocom},
  \bibinfo{author}{J.~D. Young}, \bibinfo{author}{J.~M. Iniguez},
  \bibinfo{title}{Climate relationships with increasing wildfire in the
  southwestern {US} from 1984 to 2015}, \bibinfo{journal}{Forest Ecology and
  Management} \bibinfo{volume}{460} (\bibinfo{year}{2020})
  \bibinfo{pages}{117861}.

\bibitem[{Thomas et~al.(2017)Thomas, Butry, Gilbert, Webb, and
  Fung}]{thomas_costs_2017}
\bibinfo{author}{D.~Thomas}, \bibinfo{author}{D.~Butry},
  \bibinfo{author}{S.~Gilbert}, \bibinfo{author}{D.~Webb},
  \bibinfo{author}{J.~Fung}, \bibinfo{title}{The costs and losses of wildfires:
  {A} literature survey}, \bibinfo{type}{NIST Special Publication 1215},
  National Institute of Standards and Technology,
  \bibinfo{address}{Gaithersburg, MD}, \bibinfo{year}{2017}.

\bibitem[{Kollanus et~al.(2017)Kollanus, Prank, Gens, Soares, Vira, Kukkonen,
  Sofiev, Salonen, and Lanki}]{kollanus_mortality_2017}
\bibinfo{author}{V.~Kollanus}, \bibinfo{author}{M.~Prank},
  \bibinfo{author}{A.~Gens}, \bibinfo{author}{J.~Soares},
  \bibinfo{author}{J.~Vira}, \bibinfo{author}{J.~Kukkonen},
  \bibinfo{author}{M.~Sofiev}, \bibinfo{author}{R.~O. Salonen},
  \bibinfo{author}{T.~Lanki}, \bibinfo{title}{Mortality due to vegetation
  fire-originated {PM2.5} exposure in {Europe}-{Assessment} for the years 2005
  and 2008}, \bibinfo{journal}{Environmental Health Perspectives}
  \bibinfo{volume}{125} (\bibinfo{year}{2017}) \bibinfo{pages}{30--37}.

\bibitem[{van~der Werf et~al.(2017)van~der Werf, Randerson, Giglio, van
  Leeuwen, Chen, Rogers, Mu, van Marle, Morton, Collatz, Yokelson, and
  Kasibhatla}]{van_der_werf_global_2017}
\bibinfo{author}{G.~R. van~der Werf}, \bibinfo{author}{J.~T. Randerson},
  \bibinfo{author}{L.~Giglio}, \bibinfo{author}{T.~T. van Leeuwen},
  \bibinfo{author}{Y.~Chen}, \bibinfo{author}{B.~M. Rogers},
  \bibinfo{author}{M.~Mu}, \bibinfo{author}{M.~J.~E. van Marle},
  \bibinfo{author}{D.~C. Morton}, \bibinfo{author}{G.~J. Collatz},
  \bibinfo{author}{R.~J. Yokelson}, \bibinfo{author}{P.~S. Kasibhatla},
  \bibinfo{title}{Global fire emissions estimates during 1997–2016},
  \bibinfo{journal}{Earth System Science Data} \bibinfo{volume}{9}
  (\bibinfo{year}{2017}) \bibinfo{pages}{697--720}.

\bibitem[{Bakhshaii and Johnson(2019)}]{BAKHSHAII_JOHNSON_ETAL_CJFR2019}
\bibinfo{author}{A.~Bakhshaii}, \bibinfo{author}{E.~A. Johnson},
  \bibinfo{title}{A review of a new generation of wildfire-atmosphere
  modeling}, \bibinfo{journal}{Canadian Journal of Forest Research}
  \bibinfo{volume}{49} (\bibinfo{year}{2019}) \bibinfo{pages}{565--574}.

\bibitem[{Sullivan(2009{\natexlab{a}})}]{Sullivan2009a}
\bibinfo{author}{A.~L. Sullivan}, \bibinfo{title}{Wildland surface fire spread
  modelling, 1990--2007. 1: {Physical} and quasi-physical models},
  \bibinfo{journal}{International Journal of Wildland Fire}
  \bibinfo{volume}{18} (\bibinfo{year}{2009}{\natexlab{a}})
  \bibinfo{pages}{349--368}.

\bibitem[{Sullivan(2009{\natexlab{b}})}]{Sullivan2009b}
\bibinfo{author}{A.~L. Sullivan}, \bibinfo{title}{Wildland surface fire spread
  modelling, 1990--2007. 2: {Empirical} and quasi-empirical models},
  \bibinfo{journal}{International Journal of Wildland Fire}
  \bibinfo{volume}{18} (\bibinfo{year}{2009}{\natexlab{b}})
  \bibinfo{pages}{369–386}.

\bibitem[{Sullivan(2009{\natexlab{c}})}]{Sullivan2009c}
\bibinfo{author}{A.~L. Sullivan}, \bibinfo{title}{Wildland surface fire spread
  modelling, 1990--2007. 3: {Simulation} and mathematical analogue models},
  \bibinfo{journal}{International Journal of Wildland Fire}
  \bibinfo{volume}{18} (\bibinfo{year}{2009}{\natexlab{c}})
  \bibinfo{pages}{387--403}.

\bibitem[{Linn and Harlow(1997)}]{Linn1997}
\bibinfo{author}{R.~R. Linn}, \bibinfo{author}{F.~H. Harlow},
  \bibinfo{title}{{FIRETEC}: {A} transport description of wildfire behavior},
  \bibinfo{type}{LA-UR-97-3920}, Los Alamos National Laboratory,
  \bibinfo{address}{Los Alamos, NM (United States)}, \bibinfo{year}{1997}.

\bibitem[{Larini et~al.(1998)Larini, Giroud, Porterie, and
  Loraud}]{larini_multiphase_1998}
\bibinfo{author}{M.~Larini}, \bibinfo{author}{F.~Giroud},
  \bibinfo{author}{B.~Porterie}, \bibinfo{author}{J.~C. Loraud},
  \bibinfo{title}{A multiphase formulation for fire propagation in
  heterogeneous combustible media}, \bibinfo{journal}{International Journal of
  Heat and Mass Transfer} \bibinfo{volume}{41} (\bibinfo{year}{1998})
  \bibinfo{pages}{881--897}.

\bibitem[{Mell et~al.(2007)Mell, Jenkins, Gould, and
  Cheney}]{mell_physics-based_2007}
\bibinfo{author}{W.~Mell}, \bibinfo{author}{M.~A. Jenkins},
  \bibinfo{author}{J.~Gould}, \bibinfo{author}{P.~Cheney}, \bibinfo{title}{A
  physics-based approach to modelling grassland fires},
  \bibinfo{journal}{International Journal of Wildland Fire}
  \bibinfo{volume}{16} (\bibinfo{year}{2007}) \bibinfo{pages}{1--22}.

\bibitem[{Rothermel(1972)}]{Rothermel1972}
\bibinfo{author}{R.~C. Rothermel}, \bibinfo{title}{A Mathematical Model for
  Predicting Fire Spread in Wildland Fuels}, \bibinfo{type}{Research Paper
  INT-115}, USDA Forest Service, \bibinfo{address}{Intermountain Forest and
  Range Experiment Station, Ogden, UT 84401}, \bibinfo{year}{1972}.

\bibitem[{Andrews(1986)}]{andrews_behave_1986}
\bibinfo{author}{P.~L. Andrews}, \bibinfo{title}{{BEHAVE}: Fire Behavior
  Prediction and Fuel Modeling system--{BURN} Subsystem, Part 1},
  \bibinfo{type}{General Technical Report INT-194}, USDA Forest Service,
  \bibinfo{address}{Intermountain Research Station, Ogden, UT 84401},
  \bibinfo{year}{1986}.

\bibitem[{Finney(1998)}]{Finney1998}
\bibinfo{author}{M.~A. Finney}, \bibinfo{title}{FARSITE: Fire Area
  Simulator--Model Development and Evaluation}, \bibinfo{type}{Research Paper
  RMRS-RP-4}, USDA Forest Service, Rocky Mountain Research Station,
  \bibinfo{year}{1998}. \bibinfo{note}{Revised 2004}.

\bibitem[{Jain et~al.(2020)Jain, Coogan, Subramanian, Crowley, Taylor, and
  Flannigan}]{JAIN_ETAL_ER2020}
\bibinfo{author}{P.~Jain}, \bibinfo{author}{S.~C.~P. Coogan},
  \bibinfo{author}{S.~G. Subramanian}, \bibinfo{author}{M.~Crowley},
  \bibinfo{author}{S.~Taylor}, \bibinfo{author}{M.~D. Flannigan},
  \bibinfo{title}{A review of machine learning applications in wildfire science
  and management}, \bibinfo{journal}{Environmental Reviews}
  \bibinfo{volume}{28} (\bibinfo{year}{2020}) \bibinfo{pages}{478--505}.

\bibitem[{Ihme et~al.(2022)Ihme, Chung, and
  Mishra}]{IHME_CHUNG_MISHRA_PECS2022}
\bibinfo{author}{M.~Ihme}, \bibinfo{author}{W.~T. Chung},
  \bibinfo{author}{A.~A. Mishra}, \bibinfo{title}{Combustion machine learning:
  {Principles, progress and prospects}}, \bibinfo{journal}{Progress in Energy
  and Combustion Science} \bibinfo{volume}{91} (\bibinfo{year}{2022})
  \bibinfo{pages}{101010}.

\bibitem[{LeCun et~al.(2015)LeCun, Bengio, and Hinton}]{LECUN_ETAL_N2015}
\bibinfo{author}{Y.~LeCun}, \bibinfo{author}{Y.~Bengio},
  \bibinfo{author}{G.~Hinton}, \bibinfo{title}{Deep learning},
  \bibinfo{journal}{Nature} \bibinfo{volume}{521} (\bibinfo{year}{2015})
  \bibinfo{pages}{436–444}.

\bibitem[{Hodges and Lattimer(2019)}]{Hodges2019}
\bibinfo{author}{J.~L. Hodges}, \bibinfo{author}{B.~Y. Lattimer},
  \bibinfo{title}{Wildland fire spread modeling using convolutional neural
  networks}, \bibinfo{journal}{Fire Technology} \bibinfo{volume}{55}
  (\bibinfo{year}{2019}) \bibinfo{pages}{2115--2142}.

\bibitem[{Radke et~al.(2019)Radke, Hessler, and Ellsworth}]{firecast}
\bibinfo{author}{D.~Radke}, \bibinfo{author}{A.~Hessler},
  \bibinfo{author}{D.~Ellsworth}, in: \bibinfo{booktitle}{Proceedings of the
  Twenty-Eighth International Joint Conference on Artificial Intelligence,
  {IJCAI-19}}, International Joint Conferences on Artificial Intelligence
  Organization, \bibinfo{year}{2019}, pp. \bibinfo{pages}{4575--4581}.

\bibitem[{Burge et~al.(2020)Burge, Bonanni, Ihme, and Hu}]{Burge2020}
\bibinfo{author}{J.~Burge}, \bibinfo{author}{M.~Bonanni},
  \bibinfo{author}{M.~Ihme}, \bibinfo{author}{L.~Hu},
  \bibinfo{title}{Convolutional {LSTM} neural networks for modeling wildland
  fire dynamics}, \bibinfo{journal}{arXiv Preprint}
  \bibinfo{volume}{2012.06679} (\bibinfo{year}{2020}).

\bibitem[{Bolt et~al.(2022)Bolt, Huston, Kuhnert, Dabrowski, Hilton, and
  Sanderson}]{Bolt2022}
\bibinfo{author}{A.~Bolt}, \bibinfo{author}{C.~Huston},
  \bibinfo{author}{P.~Kuhnert}, \bibinfo{author}{J.~J. Dabrowski},
  \bibinfo{author}{J.~Hilton}, \bibinfo{author}{C.~Sanderson},
  \bibinfo{title}{A spatio-temporal neural network forecasting approach for
  emulation of firefront models}, \bibinfo{journal}{arXiv Preprint}
  \bibinfo{volume}{2206.08523} (\bibinfo{year}{2022}).

\bibitem[{Kulkarni et~al.(2015)Kulkarni, Whitney, Kohli, and
  Tenenbaum}]{Kulkarni2015}
\bibinfo{author}{T.~D. Kulkarni}, \bibinfo{author}{W.~F. Whitney},
  \bibinfo{author}{P.~Kohli}, \bibinfo{author}{J.~Tenenbaum}, in:
  \bibinfo{editor}{C.~Cortes}, \bibinfo{editor}{N.~Lawrence},
  \bibinfo{editor}{D.~Lee}, \bibinfo{editor}{M.~Sugiyama},
  \bibinfo{editor}{R.~Garnett} (Eds.), \bibinfo{booktitle}{Advances in Neural
  Information Processing Systems}, volume~\bibinfo{volume}{28}, Curran,
  \bibinfo{year}{2015}.

\bibitem[{Sanchez-Gonzalez et~al.(2020)Sanchez-Gonzalez, Godwin, Pfaff, Ying,
  Leskovec, and Battaglia}]{EPD}
\bibinfo{author}{A.~Sanchez-Gonzalez}, \bibinfo{author}{J.~Godwin},
  \bibinfo{author}{T.~Pfaff}, \bibinfo{author}{R.~Ying},
  \bibinfo{author}{J.~Leskovec}, \bibinfo{author}{P.~Battaglia}, in:
  \bibinfo{editor}{H.~Daum\'{e}~III}, \bibinfo{editor}{A.~Singh} (Eds.),
  \bibinfo{booktitle}{Proceedings of the 37th International Conference on
  Machine Learning}, volume \bibinfo{volume}{119} of
  \textit{\bibinfo{series}{Proceedings of Machine Learning Research}}, pp.
  \bibinfo{pages}{8459--8468}.

\bibitem[{Kochkov et~al.(2021)Kochkov, Smith, Alieva, Wang, Brenner, and
  Hoyer}]{EPD_stephan}
\bibinfo{author}{D.~Kochkov}, \bibinfo{author}{J.~A. Smith},
  \bibinfo{author}{A.~Alieva}, \bibinfo{author}{Q.~Wang},
  \bibinfo{author}{M.~P. Brenner}, \bibinfo{author}{S.~Hoyer},
  \bibinfo{title}{Machine learning accelerated computational fluid dynamics},
  \bibinfo{journal}{Proceedings of the National Academy of Sciences}
  \bibinfo{volume}{118} (\bibinfo{year}{2021}) \bibinfo{pages}{e2101784118}.

\bibitem[{Scott and Burgan(2005)}]{Scott2005}
\bibinfo{author}{J.~H. Scott}, \bibinfo{author}{R.~E. Burgan},
  \bibinfo{title}{Standard Fire Behavior Fuel Models: {A} Comprehensive Set for
  Use with {Rothermel's} Surface Fire Spread Model}, \bibinfo{type}{General
  Technical Report RMRS-GTR-153}, USDA Forest Service Rocky Mountain Research
  Station, \bibinfo{year}{2005}.

\bibitem[{Rollins and Rollins(2009)}]{Rollins2009}
\bibinfo{author}{M.~G. Rollins}, \bibinfo{author}{M.~G. Rollins},
  \bibinfo{title}{{LANDFIRE}: {A} nationally consistent vegetation, wildland
  fire, and fuel assessment}, \bibinfo{journal}{International Journal of
  Wildland Fire} \bibinfo{volume}{18} (\bibinfo{year}{2009})
  \bibinfo{pages}{235--249}.

\bibitem[{Foster(2019)}]{GenerativeDeepLearning}
\bibinfo{author}{D.~Foster}, \bibinfo{title}{Generative Deep Learning},
  O'Reilly, \bibinfo{year}{2019}.

\bibitem[{He et~al.(2016)He, Zhang, Ren, and Sun}]{resnet}
\bibinfo{author}{K.~He}, \bibinfo{author}{X.~Zhang}, \bibinfo{author}{S.~Ren},
  \bibinfo{author}{J.~Sun}, in: \bibinfo{booktitle}{IEEE Conference on Computer
  Vision and Pattern Recognition (CVPR)}, \bibinfo{address}{Las Vegas, NV}, pp.
  \bibinfo{pages}{770--778}.

\bibitem[{Abadi et~al.(2015)Abadi, Agarwal, Barham, Brevdo, Chen, Citro,
  Corrado, Davis, Dean, Devin, Ghemawat, Goodfellow, Harp, Irving, Isard, Y.,
  Jozefowicz, Kaiser, Kudlur, Levenberg, Man\'{e}, Monga, Moore, Murray, Olah,
  Schuster, Shlens, Steiner, Sutskever, Talwar, Tucker, Vanhoucke, Vasudevan,
  Vi\'{e}gas, Vinyals, Warden, Wattenberg, Wicke, Yu, and Zheng}]{tensorflow}
\bibinfo{author}{M.~Abadi}, \bibinfo{author}{A.~Agarwal},
  \bibinfo{author}{P.~Barham}, \bibinfo{author}{E.~Brevdo},
  \bibinfo{author}{Z.~Chen}, \bibinfo{author}{C.~Citro}, \bibinfo{author}{G.~S.
  Corrado}, \bibinfo{author}{A.~Davis}, \bibinfo{author}{J.~Dean},
  \bibinfo{author}{M.~Devin}, \bibinfo{author}{S.~Ghemawat},
  \bibinfo{author}{I.~Goodfellow}, \bibinfo{author}{A.~Harp},
  \bibinfo{author}{G.~Irving}, \bibinfo{author}{M.~Isard},
  \bibinfo{author}{Y.}, \bibinfo{author}{R.~Jozefowicz},
  \bibinfo{author}{L.~Kaiser}, \bibinfo{author}{M.~Kudlur},
  \bibinfo{author}{J.~Levenberg}, \bibinfo{author}{D.~Man\'{e}},
  \bibinfo{author}{R.~Monga}, \bibinfo{author}{S.~Moore},
  \bibinfo{author}{D.~Murray}, \bibinfo{author}{C.~Olah},
  \bibinfo{author}{M.~Schuster}, \bibinfo{author}{J.~Shlens},
  \bibinfo{author}{B.~Steiner}, \bibinfo{author}{I.~Sutskever},
  \bibinfo{author}{K.~Talwar}, \bibinfo{author}{P.~Tucker},
  \bibinfo{author}{V.~Vanhoucke}, \bibinfo{author}{V.~Vasudevan},
  \bibinfo{author}{F.~Vi\'{e}gas}, \bibinfo{author}{O.~Vinyals},
  \bibinfo{author}{P.~Warden}, \bibinfo{author}{M.~Wattenberg},
  \bibinfo{author}{M.~Wicke}, \bibinfo{author}{Y.~Yu},
  \bibinfo{author}{X.~Zheng}, \bibinfo{title}{{TensorFlow}: Large-scale machine
  learning on heterogeneous systems}, \bibinfo{year}{2015}.
  \bibinfo{note}{Software available from \url{https://www.tensorflow.org}}.

\bibitem[{Chollet et~al.(2015)}]{keras}
\bibinfo{author}{F.~Chollet}, et~al., \bibinfo{title}{Keras},
  \bibinfo{howpublished}{\url{https://keras.io}}, \bibinfo{year}{2015}.

\bibitem[{Kingma and Ba(2015)}]{adam_optimizer}
\bibinfo{author}{D.~P. Kingma}, \bibinfo{author}{J.~Ba}, in:
  \bibinfo{booktitle}{Proceedings of the International Conference for Learning
  Representations}.

\bibitem[{Filippi et~al.(2014)Filippi, Mallet, and Nader}]{shape_agreement}
\bibinfo{author}{J.-B. Filippi}, \bibinfo{author}{V.~Mallet},
  \bibinfo{author}{B.~Nader}, \bibinfo{title}{Representation and evaluation of
  wildfire propagation simulations}, \bibinfo{journal}{International Journal of
  Wildland Fire} \bibinfo{volume}{23} (\bibinfo{year}{2014})
  \bibinfo{pages}{46}.

\bibitem[{Tibshirani and Efron(1993)}]{tibshirani_1993}
\bibinfo{author}{R.~J. Tibshirani}, \bibinfo{author}{B.~Efron},
  \bibinfo{title}{An Introduction to the Bootstrap},
  volume~\bibinfo{volume}{57} of \textit{\bibinfo{series}{Monographs on
  Statistics and Applied Probability}}, Chapman and Hall, \bibinfo{year}{1993}.

\bibitem[{Rombach et~al.(2021)Rombach, Blattmann, Lorenz, Esser, and
  Ommer}]{stable_diffusion}
\bibinfo{author}{R.~Rombach}, \bibinfo{author}{A.~Blattmann},
  \bibinfo{author}{D.~Lorenz}, \bibinfo{author}{P.~Esser},
  \bibinfo{author}{B.~Ommer}, \bibinfo{title}{High-resolution image synthesis
  with latent diffusion models}, \bibinfo{year}{2021}.

\bibitem[{Shi et~al.(2015)Shi, Chen, Wang, Yeung, Wong, and Woo}]{ConvLSTM}
\bibinfo{author}{X.~Shi}, \bibinfo{author}{Z.~Chen}, \bibinfo{author}{H.~Wang},
  \bibinfo{author}{D.-Y. Yeung}, \bibinfo{author}{W.-k. Wong},
  \bibinfo{author}{W.-c. Woo}, in: \bibinfo{editor}{C.~Cortes},
  \bibinfo{editor}{N.~Lawrence}, \bibinfo{editor}{D.~Lee},
  \bibinfo{editor}{M.~Sugiyama}, \bibinfo{editor}{R.~Garnett} (Eds.),
  \bibinfo{booktitle}{Advances in Neural Information Processing Systems},
  volume~\bibinfo{volume}{28}, Curran, \bibinfo{year}{2015}.

\end{thebibliography}

\appendix
%\begin{appendices}
\section{Deep Neural Network Transformations}
\label{appendix:brief dnn background}
While a full summary of deep neural networks is outside the scope of this paper, we provide a brief description of the transformations used in the DNNs described in this work.  For a deeper background on DNNs, we refer the interested reader to \cite{GenerativeDeepLearning}.

The fundamental unit processed by a DNN is a tensor.  A tensor is an $n$ dimensional collection of scalar values, $n\geq0$.  For example, a single field specifying how much fuel exists in a 2D patch of ground could be stored in a 2D tensor with shape $(H, W)$ where $H$ is the height of the field and $W$ is the width.  That field represents a \textit{channel} of information and if there are additional channels, they can be stored in a tensor of shape $(H, W, C)$ where $C$ is the number of channels.  A time series of fields can be stored in a tensor of shape $(T, H, W, C)$ where $T$ is the number time points in the series.  Training a DNN is usually done in batches, so a DNN that works on $(T, H, W, C)$ data will actually take $(B, T, H, W, C)$ data where $B$ is the number of data points in a single batch.

DNNs are often said to \textit{predict} some outcome based on a given data point though in reality, the output is merely the input after its been passed through a potentially large number of transformations.  We sometimes refer to sets of transformations as a \textit{layer}.  Some layers contain parameters such that the process of training a DNN attempts to find the optimal set of values that result in transforming as much of the training input into as correct a set of output as possible.  These are the primary transformations used in this work:

\begin{itemize}
    \item[] \texttt{fully connected}: Every value in the output is a parameterized combination of every value in the input.  Given an input shape $(B, M)$ and output shape $(B, N)$, there are $M*N$ total parameters. This layer can be useful when many output values need to consider many input values, but can be particularly prone to overfitting and parameter bloat.
    \item[] \texttt{convolutional (2D)}: This layer is appropriate for inputs that contain 2D fields.  A 2D convolutional kernel is swept across both dimensions of the input field.  At each location, the dot product between the kernel and the 2D field is computed, which generates a new field that is the convolution of the input field with the kernel.  There can be multiple independent kernels, resulting in multiple field outputs.  Given an input shape of $(B, H, W, C)$, the output shape is $(B, H, W, K)$ where $K$ is the number of kernels.
    \item[] \texttt{recurrent}:  A layer which explicitly considers temporal relationships in the input data when generating the output.  Unlike spatial relationships, temporal relationships often cannot effectively be modeled by simply considering neighboring input. Temporal dynamics often require considering events that occurred at more distant times in the past.  Recurrent layers build up a \textit{memory} by iterating over individual time points one at a time.  Cells in the output depend on the memory instead of actual time points in the past.  If the input contains fields, the layers will also leverage convolutions such that an output cell can depend on the memory of itself, and the memory of neighboring cells.  If the convolutions have $K$ kernels and the input shape is $(B, T, H, W, C)$, then the output shape will be $(B, T, H, W, K)$.  Given its wide-spread success at modeling image-to-image type tasks, we use the Convolutional Long-Short Term Memory layer (ConvLSTM) \cite{ConvLSTM}.
    \item[] \texttt{activation}: Passes each value in the input through an often non-linear and potentially parameterized function.  The output shape is equal to the input shape.
    \item[] \texttt{max pooling (2D)}: A field transformation that downsamples the spatial dimensions of the input and while typically doubling the number of channels.  If the input has shape $(B, H, W, C)$, then the output will have shape $(B, H/2, W/2, C*2)$.
    \item[] \texttt{zero padding}: This layer adds padding around the field.  If the input shape is $(B, H, W, C)$ and the padding size is $P$ then the output shape will be $(B, H+2P, W+2P, C)$.
    \item[] \texttt{one hot encoding}: This layer transforms integer values in the input into a list of all zeros with the exception of a single element in the list corresponding to the integer value being given a value of one.  If the input shape is $(B, H, W, C)$, the output will be $(B, H, W, C, E)$ where $E$ is the maximum value the input can take.
    \item[] \texttt{embedding}: This layer converts a field that contains discrete valued cells into an field where each cell is replaced by a set of floating-point values.  Values in the input that result in similar predictions are placed close to each other in the embedding space, which subsequent layers in the model can leverage more effectively than the discrete values alone.  If the input has shape $(B, H, W, 1)$ and the embedding space has rank $E$, then the output shape will be $(B, H, W, E)$.
\end{itemize}

%\end{appendices}

\end{document}